\documentclass{article}

\usepackage[preprint]{corl_2026} % Uncomment for the camera-ready ``final'' version.

\usepackage{graphicx}
\usepackage{amsmath, amssymb}
\usepackage{xspace}
\usepackage{booktabs, multirow,tabularx, array}
\usepackage{xcolor}
\usepackage{wrapfig}
\usepackage{booktabs}
\usepackage{float}
\usepackage{makecell}

\newcommand{\model}{GeomVLA}

\newcommand*{\etc}{%
    \@ifnextchar{.}%
        {\textit{etc}}%
        {\textit{etc.}\@\xspace}%
}
\title{\model{}: Unifying Scene, Motion, and Action in 3D}

\author{
  Ziyin Xiong$^{1}$ \quad Nikolaos Gkanatsios$^{1}$ \quad Moritz Reuss$^{2}$ \quad Katerina Fragkiadaki$^{1}$\\
  \normalfont $^{1}$Carnegie Mellon University \quad $^{2}$NVIDIA\\
  \normalfont \texttt{\{ziyinx,katef\}@andrew.cmu.edu}
}

\begin{document}
\maketitle

%===============================================================================

\begin{abstract}
We present \textbf{\model{}}, a Vision-Language-Action (VLA) model that unifies perception, latent scene motion prediction, and action generation within a shared robot-centric 3D coordinate frame. Our approach lifts pretrained VLM features into spatially grounded 3D scene tokens using depth and camera calibration, while retaining the semantic representations learned during VLM pretraining. We further introduce a \emph{3D Scene Trajectory Denoiser}, a task-conditioned module that learns a latent representation of how scene points are expected to move in 3D. Rather than executing the predicted trajectory as an open-loop plan, \model{} extracts intermediate motion tokens from the trajectory denoiser and uses them to condition a 3D flow-based action denoiser through geometry-aware attention. \model{} achieves state-of-the-art performance on CALVIN, competitive performance on LIBERO and RoboTwin2.0, and outperforms strong baselines in real-world manipulation settings without robot-action pretraining. Extensive ablations show that future-motion reasoning alone is insufficient: the primary gains are associated with maintaining geometric consistency among scene representation, motion prediction, and robot actions throughout the perception-to-action pipeline.
Project webpage: \href{https://ziyin-xiong.github.io/geomvla.io/}{ziyin-xiong.github.io/geomvla.io}.
\end{abstract}


\begin{figure}[H]
    \centering
    \includegraphics[width=\linewidth]{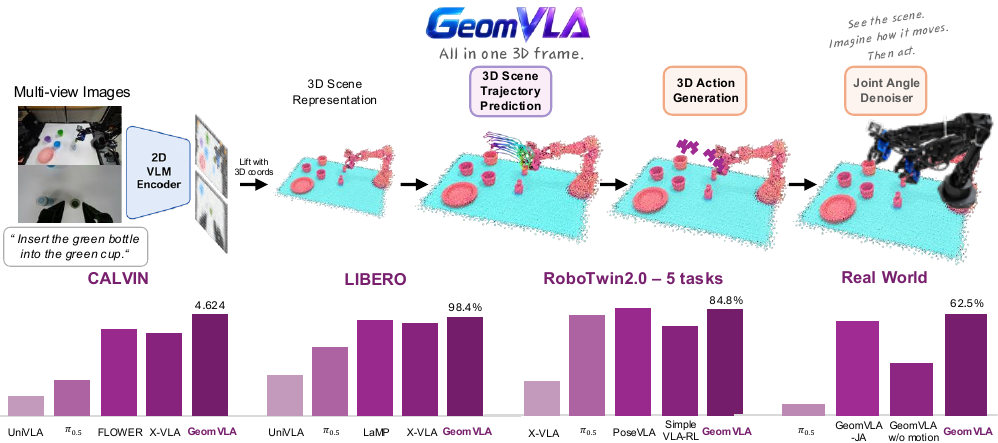}
    \caption{\textbf{GeomVLA enables motion-guided 3D visual-language-action learning by bridging perception, latent future motion, and robot control in a shared 3D space.} It lifts pretrained 2D VLM features into a geometrically grounded 3D representation, predicts latent 3D scene trajectories as an intermediate reasoning signal, and conditions a 3D action denoiser on the resulting scene-motion representation to generate robot actions. This design improves performance in both simulation and the real world. It also supports joint-angle action denoising by conditioning on scene-motion and end-effector pose-action features.}
    \label{fig:teaser}
\end{figure}

%===============================================================================

\section{Introduction}
\label{sec:introduction}

Vision-language-action (VLA) models \cite{brohan2023rt2visionlanguageactionmodelstransfer,black2026pi0visionlanguageactionflowmodel,intelligence2025pi05visionlanguageactionmodelopenworld} have recently emerged as a promising paradigm for language-conditioned robot manipulation by leveraging the broad semantic knowledge of pretrained vision-language models (VLMs) \cite{beyer2024paligemmaversatile3bvlm,qwen2025qwen25technicalreport,xiao2023florence2advancingunifiedrepresentation}. However, most existing VLAs operate primarily on 2D images while robot manipulation fundamentally requires reasoning about geometry, contact, and motion in 3D space \cite{reuss2025flowerdemocratizinggeneralistrobot,zheng2025xvlasoftpromptedtransformerscalable,intelligence2025pi05visionlanguageactionmodelopenworld}. This creates a mismatch between image-centric representations and the metric geometry required for control.

One natural solution is to represent both the scene and robot actions directly in metric 3D coordinates. Recent 3D manipulation policies show that explicit geometric representations improve spatial reasoning and data efficiency~\cite{gervet2023act3d,shridhar2023perceiver,ke20243d,goyal2023rvt,yang2025fp33dfoundationpolicy}. More recent VLA systems combine such geometric representations with pretrained VLMs through lifted 3D features, spatial positional encodings, or dual-stream architectures~\cite{zhen20243dvla3dvisionlanguageactiongenerative,qu2025spatialvlaexploringspatialrepresentations,zhang20254dvlaspatiotemporalvisionlanguageactionpretraining,jia2024lift3dfoundationpolicylifting,li2025pointvlainjecting3dworld,sun2026geovla}. Although these approaches improve geometric grounding, they primarily focus on static 3D scene representations and action generation, without addressing how intermediate future reasoning should be represented and coupled to control.

A complementary line of work introduces future prediction as an intermediate reasoning process for robot manipulation~\cite{cen2025worldvlaautoregressiveactionworld,zhong2025flowvlavisualchainthoughtbased,zheng2025tracevlavisualtraceprompting,gu2023rttrajectoryrobotictaskgeneralization,lee2025tracegenworldmodeling3d,lin2026roboflow4dlightweightflowworld,zhi20253dflowactionlearningcrossembodimentmanipulation}. Instead of mapping observations directly to actions, these methods encourage policies to reason about how the scene will evolve through predicted videos, optical flow, visual traces, or trajectories. Such representations can expose task dynamics and support long-horizon planning. Many of these methods perform future reasoning in pixel or image space, whereas actions are executed in 3D. We argue that considering 3D representation and future prediction separately leaves a mismatch among scene understanding, motion reasoning, and control (Figure~\ref{fig:teaser}). Consequently, these components do not interact through a common geometric representation.

In this work, we introduce \textbf{\model{}}, a 3D motion-aware VLA that unifies perception, latent future-motion reasoning, and action generation within a shared robot-centric 3D coordinate frame. We hypothesize that future-motion reasoning is most useful when perception, motion prediction, and control share the same geometry. \model{} therefore expresses scene features, predicted motion, and Cartesian actions in a common robot-centric metric frame. Concretely, \model{} first lifts pretrained VLM features into spatially grounded 3D scene tokens using depth and camera calibration. We then introduce a \emph{3D Scene Trajectory Denoiser}, a task-conditioned module that predicts future 3D point motion and exposes intermediate motion tokens to the action policy. Rather than executing a predicted trajectory directly, the action policy uses intermediate trajectory-denoiser features as a latent geometric reasoning signal. Consequently, scene features, motion representations, and robot trajectories interact within a unified metric coordinate frame throughout action generation.

We evaluate \model{} across CALVIN~\cite{mees2022calvin}, LIBERO~\cite{liu2024libero}, RoboTwin2.0~\cite{chen2025robotwin20scalabledata}, and real-world manipulation benchmarks. Our approach achieves state-of-the-art performance on CALVIN, competitive performance on LIBERO and RoboTwin2.0, and outperforms strong baselines in real-world settings without requiring robot-action pretraining. Extensive ablations show that future-motion reasoning alone is insufficient: the primary gains are associated with maintaining geometric consistency between scene representation, motion prediction, and robot actions throughout the perception-to-action pipeline.

In summary, our contributions are as follows:
\begin{itemize}
    \item We introduce GeomVLA, a VLA framework that unifies scene representation, latent future-motion reasoning, and action generation within a shared 3D coordinate frame.

    \item We propose a \emph{3D Scene Trajectory Denoiser}, a task-conditioned module that predicts future scene motion directly in 3D space and provides intermediate motion features to downstream action generation through geometry-aware attention.

    \item We achieve state-of-the-art performance on CALVIN, competitive performance on LIBERO and RoboTwin2.0, and strong real-world results without robot-action pretraining.

    \item Extensive ablations show that maintaining geometric consistency between scene representation, motion prediction, and robot actions benefits long-horizon manipulation.
\end{itemize}

\section{Related Work}
\label{sec:related_work}

\textbf{3D Vision-Language-Action Models.}
Recent 3D manipulation policies \cite{shridhar2023perceiver,gervet2023act3d,xian2023chaineddiffuser,ke20243d,chisari2024learning,Liu2024VoxActBVA,grotz2024peract2,gkanatsios20253dflowmatchactorunified} improve spatial reasoning and data efficiency by representing scene observations and robot actions within a shared geometric space. More recent 3D VLA models \cite{zhen20243dvla3dvisionlanguageactiongenerative,qu2025spatialvlaexploringspatialrepresentations,li2025pointvlainjecting3dworld,jia2024lift3dfoundationpolicylifting,li2025bridgevlainputoutputalignmentefficient,lin2025evo0visionlanguageactionmodelimplicit,deng2025stereovlaenhancingvisionlanguageactionmodels,fan2026any3dvlaenhancingvlarobustness,li2026consisvla4dadvancingspatiotemporalconsistency} further incorporate pretrained VLM semantics into such frameworks. Existing approaches generally either inject 3D representations directly into pretrained VLM backbones through 3D positional encodings or couple VLM features with separate 3D encoders for action prediction. While these designs improve geometric grounding, they primarily focus on aligning scene representation and action generation, without explicitly addressing how intermediate reasoning processes should be represented within the control pipeline. In contrast, our method unifies scene representation, future-motion reasoning, and action generation within the same 3D coordinate frame.

\textbf{Motion Priors for Visuomotor Policies.}
Recent visuomotor policies increasingly use future prediction as an auxiliary learning signal. Prior work explores video prediction \cite{du2023learninguniversalpoliciestextguided,li2025unifiedvideoactionmodel,cen2025worldvlaautoregressiveactionworld,bi2025motusunifiedlatentaction,gigabrainteam2026gigabrain05mvlalearnsworld}, optical flow or visual traces \cite{xu2024flowcrossdomainmanipulationinterface,zhong2025flowvlavisualchainthoughtbased,zhi20253dflowactionlearningcrossembodimentmanipulation,zheng2025tracevlavisualtraceprompting,lee2025tracegenworldmodeling3d,noh20253dflowdiffusionpolicy,wang2026lamplearningvisionlanguageactionpolicies,lin2026roboflow4dlightweightflowworld,vosylius2024renderdiffusealigningimage}, object/gripper trajectories \cite{gu2023rttrajectoryrobotictaskgeneralization,bharadhwaj2024track2actpredictingpointtracks,wen2024anypointtrajectorymodelingpolicy,zawalski2025roboticcontrolembodiedchainofthought,li2025hamsterhierarchicalactionmodels,hsu2025spot}, and depth \cite{lee2025molmoactactionreasoningmodels,li2025qdepthvlaquantizeddepthprediction} to provide temporal cues beyond direct action supervision. These methods differ primarily in how future motion is represented and how it interacts with downstream control. Some approaches use explicit plan-then-act formulations \cite{wen2024anypointtrajectorymodelingpolicy,zheng2025tracevlavisualtraceprompting}, where predicted trajectories or flow fields are generated before policy execution, while others incorporate motion prediction as a latent supervisory signal within the policy itself \cite{wang2026lamplearningvisionlanguageactionpolicies}. These approaches thus predominantly reason in pixel or image-space representations, even when actions are ultimately executed in 3D. In contrast, our method represents future scene motion directly within the same robot-centric 3D coordinate frame used for scene representation and action generation.

\section{Method}
\label{sec:method}

\begin{figure*}[t!]
    \centering
    \includegraphics[width=\linewidth]{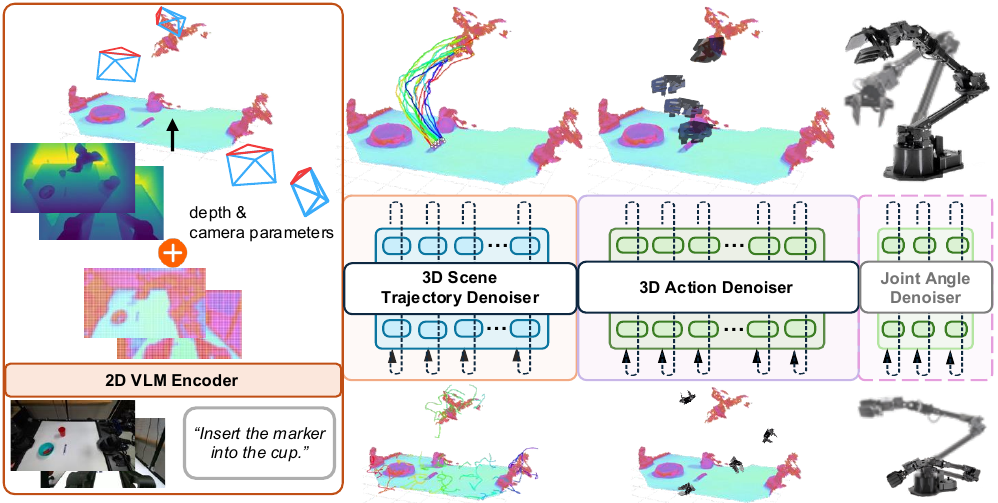}
    \caption{\textbf{\model{} architecture.}
Left: A shared 2D VLM encodes multi-view RGB observations and language instructions. Depth and camera geometry lift image features into 3D scene tokens with positions in the robot base frame. Middle: The 3D Scene Trajectory Denoiser is trained to predict future trajectories for front-view 3D anchors. During policy inference, one velocity-field evaluation provides intermediate motion features that are fused with the 3D scene tokens. Right: The 3D Action Denoiser predicts end-effector trajectory chunks conditioned on the fused scene-motion representation, language features, and robot proprioception. Optionally, the Joint Angle Denoiser uses the scene-motion representation and end-effector action features to predict joint-angle commands.
}
    \label{fig:model_arch}
\end{figure*}

Figure~\ref{fig:model_arch} illustrates \model{}. Given multi-view RGB-D observations $\{(I_v,D_v)\}_{v=1}^{V}$, a language instruction $\ell$, and robot proprioception $\mathbf{p}$, the model lifts VLM-contextualized features into 3D using depth, camera calibration, and 3D positional encoding (\S\ref{sec:scene}). A 3D Scene Trajectory Denoiser then learns an instruction-conditioned latent representation of future point motion (\S\ref{sec:actionable_3d_traj_predictor}). Its intermediate motion tokens condition a flow-based 3D action denoiser through geometry-aware attention, producing an action chunk $A=(a_1,\ldots,a_T)$ (\S\ref{sec:3d_action_policy}). Scene positions, predicted point motion, and Cartesian action tokens are all expressed in the robot base frame.

% Throughout, $\pi(\cdot)$ denotes a 3D position attached to a token; all attention operates over these positions via rotary 3D embeddings. 

\subsection{3D Scene Features}
\label{sec:scene}
A pretrained VLM~\cite{xiao2023florence2advancingunifiedrepresentation} encodes RGB observations and the language instruction $\ell$ into patch-level visual tokens $f_v$ and language embeddings $f_{\ell}$. Using depth and camera calibration, each visual token is associated with a 3D position in the robot base frame, producing lifted scene tokens $f_{v}^{\mathrm{lift}}$. Proprioceptive features $\mathbf{p}$ are represented as learnable tokens anchored at the current end-effector pose.

\subsection{3D Scene Trajectory Denoiser}
\label{sec:actionable_3d_traj_predictor}
We introduce a \emph{3D Scene Trajectory Denoiser} that predicts future point motion in the robot base frame. Intermediate decoder states provide the latent motion representation used by the downstream action denoiser. Given the front-camera RGB-D observation $(I,D)$ and camera calibration, we sample a regular grid of image points and lift them into 3D point anchors $\{q_i^{(0)}\}_{i=1}^{N}$ in the robot base frame. Our motion decoder predicts future 3D point trajectories as per-step 3D displacement increments:
\begin{equation}
\mathcal{F} = \{\Delta q_i^{(t)}\}_{i=1,t=1}^{N,H}
\in \mathbb{R}^{N \times H \times 3},
\end{equation}
where $\Delta q_i^{(t)}$ denotes the predicted displacement increment of point anchor $i$ at time step $t$. Future 3D point positions are recovered through cumulative integration $q_i^{(t)} = q_i^{(0)} + \sum_{s=1}^{t} \Delta q_i^{(s)}$.

Using the VLM features from \S\ref{sec:scene}, we bilinearly sample visual descriptors at the query locations and combine them with projected 3D coordinates. Local geometry-aware attention then produces task-conditioned 3D scene tokens $\{c_i\}_{i=1}^{N}$.

Conditioned on $\mathcal{C}=(\{c_i\},V_g,E)$, where $V_g$ and $E$ denote global visual and language tokens, a lightweight flow-matching decoder jointly predicts displacement trajectories for all anchors and future time steps. The decoder alternates serialized spatial attention over 3D anchors with temporal attention over future steps. Full encoder, decoder, and loss-weight details are provided in the appendix.

The 3D Scene Trajectory Denoiser is trained using SpatialTrackerV2 pseudo-labels extracted from benchmark demonstration videos. We supervise future displacement trajectories in the robot base frame using a motion-weighted rectified-flow objective:
\begin{equation}
\mathcal{L}_{\mathrm{traj}}
=
\mathbb{E}_{\tau,\mathcal{F}_0}
\left[
\sum_{i=1}^{N}
\sum_{t=1}^{H}
w_{i,t}
\left\|
u_{\theta}(\mathcal{F}_{\tau}, \tau, \mathcal{C})_{i,t}
-
u^{\star}_{i,t}
\right\|_2^2
\right],
\end{equation}
where $\tau$ is the flow-matching time, $\mathcal{F}_0$ a Gaussian noise trajectory, $\mathcal{F}_{\tau}$ the interpolated trajectory, $\mathcal{C}$ the visual-language context, $u_{\theta}$ the predicted velocity field, $u^{\star}_{i,t}$ the target flow at anchor $i$ and step $t$, and $w_{i,t}$ upweights motion-relevant points while retaining static structure.

\subsection{3D Action Denoiser}
\label{sec:3d_action_policy}

The 3D action denoiser predicts end-effector trajectories from lifted scene tokens $f_v^{\mathrm{lift}}$, language tokens $f_{\ell}$, proprioception $\mathbf{p}$, and the latent motion features described in \S\ref{sec:actionable_3d_traj_predictor}.

At the initial flow state $\tau=0$, we run the trajectory denoiser once on Gaussian noise, without integrating a complete trajectory. Its intermediate decoder tensor contains one feature for each of the $N=400$ anchors and $H$ future time steps. A learned temporal-attention query pools the $H$ features for each anchor into one motion token $z_i^{\mathrm{m}}$. A single geometry-aware cross-attention layer then fuses these motion tokens into the scene tokens, with the residual contribution scaled by a zero-initialized learned scalar, yielding motion-aware scene tokens $f_v^{\mathrm{mot}}$ before action generation.

The 3D action denoiser represents noisy end-effector trajectories as spatial action tokens with explicit 3D positions. A stack of geometry-aware transformer blocks alternates cross-attention to scene and language tokens with self-attention over action trajectories to produce the action feature $f_a$:
\begin{align}
f_{\mathrm{a}} &\leftarrow \mathrm{CrossAttn}_{3\mathrm{D}}\!\bigl(f_{\mathrm{a}}, [f_v^{\mathrm{mot}};f_{\ell}]\bigr), \\
f_{\mathrm{a}} &\leftarrow \mathrm{SelfAttn}_{3\mathrm{D}}(f_{\mathrm{a}}), \\
f_{\mathrm{a}} &\leftarrow \mathrm{adaLN\text{-}FFN}(f_{\mathrm{a}};\lambda,\mathbf{p}),
\end{align}
where $\lambda$ is the flow time. All attention layers use 3D rotary positional embeddings derived from token positions.

The 3D action denoiser predicts rectified-flow velocities for end-effector translation and rotation, together with binary gripper-state predictions. We optimize
\begin{equation}
\label{eq:action_loss}
\mathcal{L}_{\pi}
=
\mathbb{E}_{\lambda,\mathcal{A}_0}
\left[
\left\|
v_{\theta}(\mathcal{A}_{\lambda},\lambda,f_a)
-
v^{\star}
\right\|_2^2
+
\mathrm{BCE}(f_{\theta}^{\mathrm{open}}, a^{\mathrm{open}})\right],
\end{equation}
where $\mathcal{A}_0$ is a Gaussian noise action, $\mathcal{A}_{\lambda}$ the interpolated action, and $v^{\star}$ the target velocity. $\mathrm{BCE}$ supervises the gripper-open prediction $f_{\theta}^{\mathrm{open}}$ against ground-truth $a^{\mathrm{open}}$.

\paragraph{3D reasoning for joint-space control.}
\model{}'s 3D perception-action pathway can also serve as an intermediate geometric reasoning process for joint-space control. Instead of executing the predicted end-effector actions directly, we condition a downstream joint-angle action denoiser on $f_{\ell}$, $f_v^{\mathrm{mot}}$, and the end-effector action feature after one velocity-field evaluation. The joint-angle denoiser is trained with the same rectified-flow objective as Eq.~\eqref{eq:action_loss}.

\subsection{Training}

Training proceeds in two stages. First, we train the 3D Scene Trajectory Denoiser from SpatialTrackerV2 point-track pseudo-labels while keeping the VLM frozen. Second, we jointly fine-tune the action denoiser, pretrained trajectory denoiser, and VLM using only the action loss in Eq.~\eqref{eq:action_loss}; the trajectory pseudo-label loss is not retained in this stage. \model{} uses the public Florence-2 initialization but no robot-action pretraining and is trained only on the benchmark demonstrations.

\paragraph{Implementation details.}
To obtain pseudo-labels for the trajectory predictor, we use SpatialTrackerV2 \citep{xiao2025spatialtrackerv23dpointtracking} to estimate
3D point trajectories from videos. 
The predictor uses a $20 \times 20$ query grid of 3D anchors. 
We temporally downsample each video by a factor of three, and the predictor estimates a horizon of 15 future time steps in the downsampled sequence, spanning 45 frames at the original frame rate.
The trajectories are represented as per-step 3D displacement increments in the robot base frame.

\section{Experiments}
\label{sec:result}

We evaluate \model{} through experiments designed to isolate the role of explicit 3D future-motion reasoning in vision-language-action control: (i) whether latent 3D scene-motion prediction improves an otherwise identical 3D action-only policy, (ii) whether a shared 3D representation is beneficial, (iii) how the method compares with recent 2D and 3D VLAs and motion-guided policies, and (iv) real-world performance across varied objects, distractors, and layouts.

\subsection{Simulation Experiments}

\begin{table}[t!]
\centering
\caption{\textbf{Comparison with prior methods on simulation benchmarks.}
\model{}-JA denotes the joint-angle action variant conditioned on 3D scene-motion features and intermediate end-effector action features; 2D-JA is a 2D VLA baseline using the same VLM, joint-angle action denoiser, and training data as \model{}-JA; \model{} w/o motion removes the 3D Scene Trajectory Denoiser while retaining the 3D action policy.
\textit{Pretrained} indicates large-scale robot-action pretraining.
For RoboTwin2.0, (a)--(e) are a five-task evaluation subset selected following SimpleVLA-RL~\citep{li2025simplevlarlscalingvlatraining}, each reported as the mean success rate across the \texttt{Easy} and \texttt{Hard} settings: (a) \texttt{beat\_block\_hammer}, (b) \texttt{pick\_dual\_bottles}, (c) \texttt{move\_can\_pot}, (d) \texttt{handover\_mic}, and (e) \texttt{stack\_bowls\_two}. Tasks (a)--(b), (c)--(d), and (e) are short-, medium-, and long-horizon tasks, with approximately 100, 200, and 300 steps, respectively. We fine-tune $\pi_{0.5}$ and train full \model{} on all 50 tasks; \model{} w/o motion, 2D-JA, and \model{}-JA are trained only on the five selected tasks. The five-task comparison therefore uses different training sets and is not a fully controlled ablation.}
\label{tab:main_results}
\resizebox{\linewidth}{!}{
\begin{tabular}{l c c c ccccc cccccc}
\toprule
\multirow{2}{*}{\textbf{Method}} 
& \multirow{2}{*}{\textbf{Pretrained}}
& \multirow{2}{*}{\textbf{Params}}
& \multicolumn{1}{c}{\textbf{CALVIN}}
& \multicolumn{5}{c}{\textbf{LIBERO}}
& \multicolumn{6}{c}{\textbf{RoboTwin2.0}} \\
\cmidrule(lr){4-4} \cmidrule(lr){5-9} \cmidrule(lr){10-15}
& & 
& \textbf{ABC\_D}
& \textbf{Spatial} 
& \textbf{Object} 
& \textbf{Goal} 
& \textbf{Long} 
& \textbf{Avg.}
& \textbf{a} 
& \textbf{b} 
& \textbf{c} 
& \textbf{d} 
& \textbf{e} 
& \textbf{Avg.} \\
\midrule

\multicolumn{15}{l}{\textit{3D VLA}} \\
3D-VLA
& Yes & -- 
& 0.707 
& -- & -- & -- & -- & -- 
& -- & -- & -- & -- & -- & -- \\

SpatialVLA
& Yes & 4B 
& -- 
& 88.2 & 89.9 & 78.6 & 55.5 & 78.1 
& -- & -- & -- & -- & -- & -- \\

4D-VLA
& Yes & 4B+ 
& -- 
& 88.9 & 95.2 & 90.0 & 79.1 & 88.6 
& -- & -- & -- & -- & -- & -- \\

PoseVLA
& Yes & 3B+ 
& -- 
& 96.5 & 98.0 & 97.1 & 92.4 & 96.0 
& \textbf{93.5} & 87.0 & 58.0 & \underline{93.0} & \textbf{96.5} & \textbf{85.6} \\

\midrule
\multicolumn{15}{l}{\textit{Video/Motion-Guided}} \\
TraceVLA
& Yes & 7B
& --
& 84.6 & 85.2 & 75.1 & 54.1 & 74.8
& -- & -- & -- & -- & -- & -- \\

mimic-video
& Yes & 2B+
& --
& 94.2 & 96.8 & 90.6 & -- & 93.9
& -- & -- & -- & -- & -- & -- \\

WorldVLA
& No & 7B
& --
& 87.6 & 96.2 & 83.4 & 60.0 & 81.8
& -- & -- & -- & -- & -- & -- \\

UniVLA
& Yes & 7B
& 3.801
& 96.5 & 96.8 & 95.6 & 92.0 & 95.2
& -- & -- & -- & -- & -- & -- \\

CoT-VLA
& Yes & 7B
& --
& 87.5 & 91.6 & 87.6 & 69.0 & 81.1
& -- & -- & -- & -- & -- & -- \\

FlowVLA
& Yes & 8.5B
& --
& 93.2 & 95.0 & 91.6 & 72.6 & 88.1
& -- & -- & -- & -- & -- & -- \\

LaMP 
& Yes & 5B 
& -- 
& \textbf{99.4} & \textbf{99.8} & 97.4 & 96.7 & 98.3 
& -- & -- & -- & -- & -- & -- \\

RoboFlow4D 
& Yes & 0.76B+ 
& -- 
& 90.2 & 97.0 & 88.4 & 75.2 & 87.7
& -- & -- & -- & -- & -- & -- \\

\midrule
\multicolumn{15}{l}{\textit{Others}} \\
$\pi_{0.5}$
& Yes & 3B+ 
& 3.964
& \underline{98.8} & 98.2 & \underline{98.0} & 92.4 & 96.8
& 82.0 & 94.0 & 90.0 & 47.0 & \underline{93.0} & 81.2 \\

OpenVLA-OFT
& Yes & 7B 
& 4.279
& 97.6 & 98.4 & 97.9 & 94.5 & 97.1
& 28.1 & 29.7 & 28.1 & 45.3 & 40.6 & 34.4 \\

FLOWER
& Yes & 0.95B 
& 4.480
& 97.2 & \underline{99.3} & 96.9 & 94.5 & 97.0
& -- & -- & -- & -- & -- & -- \\

SimpleVLA-RL
& Yes & 7B 
&  --
& \textbf{99.4} & 99.1 & \textbf{99.2} & \textbf{98.5} & \textbf{99.1}
& 87.5 & 68.3 & 61.2 & 89.2 & 75.8 & 76.4 \\

X-VLA
& Yes & 0.9B 
& 4.430
& 98.2 & 98.6 & 97.8 & \underline{97.6} & 98.1
& 48.0 & 28.5 & 47.0 & 69.0 & 46.5 & 47.8 \\

\midrule

2D-JA
& No & 1B
& 4.195
& -- & -- & -- & -- & --
& 39.0 & 77.0 & 88.0 & 74.0 & 53.0 & 66.2 \\

\model{}-JA
& No & 1.3B
& \underline{4.508}
& -- & -- & -- & -- & --
& 71.0 & 75.0 & \underline{91.0} & \textbf{95.0} & 76.0 & 81.6 \\

\model{} w/o motion
& No & 1.2B
& \underline{4.508} 
& 94.8 & 99.2 & 88.2 & 91.8 & 93.5 
& 45.0 & \underline{96.0} & 83.0 & 83.0 & 43.0 & 70.0 \\

\model
& No & 1.2B
& \textbf{4.624} 
& 98.6 & \textbf{99.8} & 97.6 & \underline{97.6} & \underline{98.4} 
& \underline{89.0} & \textbf{98.0} & \textbf{99.0} & 55.0 & 83.0 & \underline{84.8} \\

\bottomrule
\end{tabular}
}
\end{table}

We evaluate \model{} on CALVIN~\citep{mees2022calvin}, LIBERO~\citep{liu2024libero}, and RoboTwin2.0~\citep{chen2025robotwin20scalabledata}, spanning long-horizon manipulation, compositional generalization, and single-/dual-arm control. The policy outputs binary gripper-state predictions and end-effector deltas for translation and rotation. For RoboTwin2.0 and the real-world tasks, every predicted pose in a chunk is a current-state-relative delta from the observed pose. For CALVIN and LIBERO, we use chained per-step deltas, with each pose defined relative to the preceding pose in the chunk.

% \paragraph{Baselines.}
% We compare 

\begin{figure*}[t!]
    \centering
    \includegraphics[width=\linewidth]{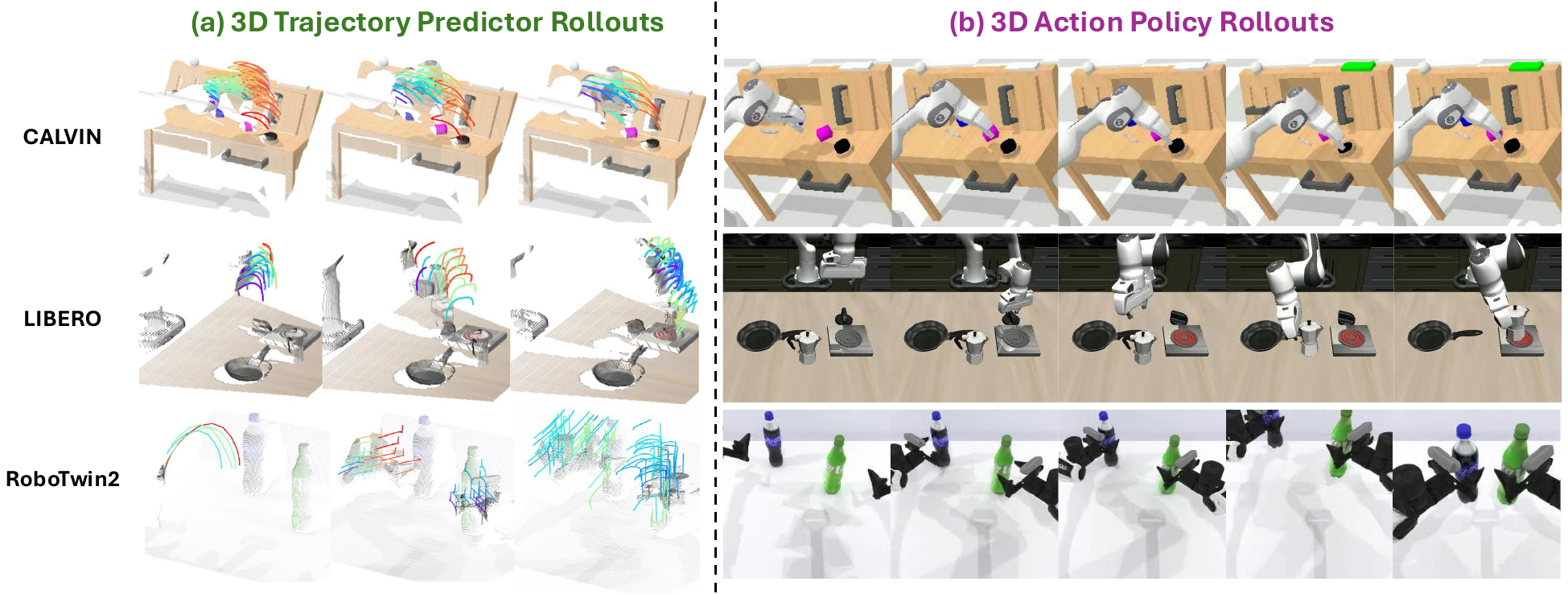}
    \caption{\textbf{Simulation examples from CALVIN, LIBERO, and RoboTwin2.0.} Left: predicted 3D point trajectories from the Scene Trajectory Denoiser. Right: corresponding closed-loop policy rollouts.}
    \label{fig:sim_results}
\end{figure*}

\paragraph{Results.} Full results are in Table~\ref{tab:main_results}, with qualitative rollouts shown in Figure~\ref{fig:sim_results}.
\model{} achieves state-of-the-art performance on CALVIN and competitive performance on LIBERO and RoboTwin2.0 against (i) 2D VLAs~\cite{intelligence2025pi05visionlanguageactionmodelopenworld,kim2025finetuningvisionlanguageactionmodelsoptimizing,reuss2025flowerdemocratizinggeneralistrobot,li2025simplevlarlscalingvlatraining,zheng2025xvlasoftpromptedtransformerscalable}, (ii) 3D policies and VLAs \cite{zhen20243dvla3dvisionlanguageactiongenerative,qu2025spatialvlaexploringspatialrepresentations,zhang20254dvlaspatiotemporalvisionlanguageactionpretraining,lin2026universalposepretraininggeneralizable}, and (iii) motion-guided VLAs \cite{zheng2025tracevlavisualtraceprompting,pai2025mimicvideovideoactionmodelsgeneralizable,cen2025worldvlaautoregressiveactionworld,li2025unifiedvideoactionmodel,zhao2025cotvlavisualchainofthoughtreasoning,zhong2025flowvlavisualchainthoughtbased,wang2026lamplearningvisionlanguageactionpolicies,lin2026roboflow4dlightweightflowworld}. On CALVIN, it improves over FLOWER (4.480 $\rightarrow$ 4.624) and X-VLA (4.430 $\rightarrow$ 4.624) without additional robot-action pretraining. On LIBERO, it attains the best average performance among methods without reinforcement-learning fine-tuning. Improvements are most pronounced in long-horizon and contact-rich tasks requiring precise spatial reasoning and delayed interaction effects. Across all 50 RoboTwin2.0 tasks, \model{} obtains 78.6\%/76.2\% success under the Easy/Hard settings, surpassing 75.9\%/75.7\% for a fully fine-tuned $\pi_{0.5}$ (Appendix Table~\ref{tab:robotwin2_all}). These results support the benefit of maintaining geometric consistency among scene representation, future reasoning, and action generation.

\subsection{Ablation Studies}
\label{sec:ablation_main}
We analyze two questions: (i) the effect of future-motion reasoning, and (ii) the role of shared 3D representation. Table~\ref{tab:ablation_results} summarizes the ablative variants of our model and their CALVIN ABC-D performance. We draw the following conclusions:

\begin{table*}[t]
\centering
\small
\setlength{\tabcolsep}{4pt}
    \caption{\textbf{CALVIN ABC-D ablations.} The metric is the mean number of consecutive tasks completed in each five-instruction sequence. The full model performs best among the evaluated variants.}
\begin{tabularx}{\textwidth}{@{}p{0.30\textwidth}Xc@{}}
\toprule
\textbf{Variant} & \textbf{Description} & \textbf{Perf.} \\
\midrule
\model{} w/o motion
& Removes the 3D Scene Trajectory Denoiser. & 4.508 \\
2DGeomVLA w/o motion
& Uses a 2D image-based policy without future-motion reasoning. & 4.411 \\
2DGeomVLA
& Uses 2D future-motion reasoning with a 2D action denoiser. & 4.437 \\
2DGeomVLA LaMP-style
& Uses pixel $(u,v)$ coordinates and metric depth for scene motion. & 4.462 \\
\model{} w/ 2D motion
& Uses image-space future motion with the 3D action denoiser. & 4.085 \\
\model{} w/ denoised trace
& Conditions action denoising on the fully denoised 3D trajectory. & 4.047 \\
\model{} w/ denoised latent
& Conditions action denoising on the fully denoised motion latent. & 4.334 \\
\model{} w/ partial denoising
& Reads the motion latent at $\tau = 0.1$ instead of pure noise. & 4.484 \\
\model{} encoder-only
& Uses only the scene trajectory encoder, without velocity prediction. & 4.443 \\
\model{} w/ arbitrary frame
& Expresses scene, motion, and action in one arbitrary rigid frame. & \underline{4.596} \\
\model{}
& Unifies scene, future motion, and action in a shared 3D frame. & \textbf{4.624} \\
\bottomrule
\end{tabularx}
\label{tab:ablation_results}
\end{table*}

\noindent \textbf{Future-motion reasoning helps} both 2D (4.411$\rightarrow$4.437) and 3D policies (4.508$\rightarrow$4.624), suggesting that predicting scene evolution provides useful temporal structure beyond action imitation.

\noindent \textbf{Explicit 3D geometry in VLAs helps.} Replacing the 2D action denoiser with the corresponding 3D action denoiser improves performance from 4.411 to 4.508, showing the benefit of representing scene features and robot actions directly in metric 3D coordinates.

\noindent \textbf{Scene motion representation matters.}
The full model conditions the action denoiser on the intermediate decoder state from one velocity-field evaluation at the initial noisy flow state. We find that weaker alternatives all underperform: using fully denoised trajectories (4.047), fully denoised latents (4.334), partially denoised latents at $\tau = 0.1$ (4.484), or only the trajectory encoder (4.443). These results suggest that the action denoiser benefits most from early-stage motion latents that capture how the scene should evolve, rather than from finalized trajectory estimates or encoder features alone.

\noindent \textbf{Geometric consistency matters.}
When future-motion reasoning is performed in image space but action generation remains in 3D, performance drops to 4.085, below the fully 2D model. The LaMP-style variant, which reasons over image-plane $(u,v)$ coordinates plus metric depth and removes 3D RoPE, reaches 4.462, suggesting that mixing image coordinates with metric depth provides a less coherent geometry for spatial reasoning. In contrast, applying the same arbitrary rigid transform to scene, motion, and action coordinates preserves performance (4.596 versus 4.624). Together, these ablations indicate that the benefit is associated with a consistent metric geometry, rather than with the robot base frame specifically. Holding the image fixed while replacing the instruction increases trajectory endpoint error by 3.40 cm on CALVIN, 5.44 cm on LIBERO, and 1.64 cm on RoboTwin2.0, showing that the predicted motion depends on language conditioning (Appendix Table~\ref{tab:traj_instruction}). Additional ablation definitions and stability analyses are provided in the appendix.

\subsection{Real-World Experiments}

\begin{figure*}[t!]
    \centering
    \includegraphics[width=\linewidth]{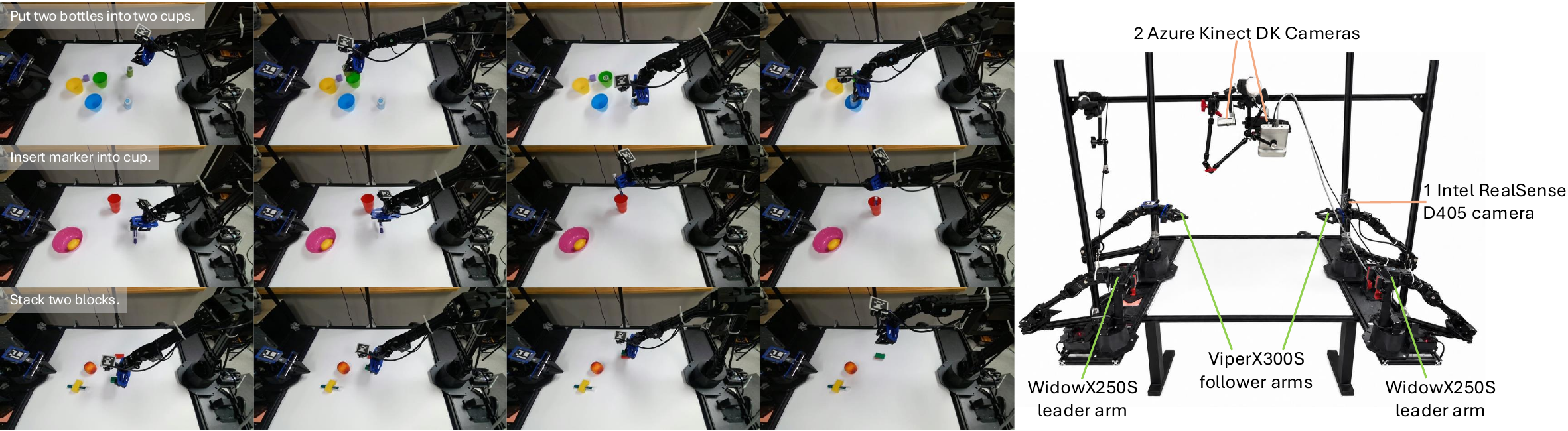}
    \caption{\textbf{Real-world evaluation setup and tasks.}
We evaluate \model{} on an ALOHA2 platform on single-arm tabletop tasks. The left panel shows representative rollouts on three real-world tasks: placing two bottles into two cups, inserting a marker into a cup, and stacking two blocks.}
    \label{fig:real-world}
\end{figure*}

\paragraph{Setup.}
We evaluate \model{} on an ALOHA2 robot platform using the right ViperX300S follower arm (Figure~\ref{fig:real-world}). Demonstrations are collected through teleoperation with the corresponding WidowX250S leader arm. Visual observations are captured by two elevated Microsoft Azure Kinect DK cameras mounted at the front and back of the workspace, together with a wrist-mounted Intel RealSense D405 camera on the right follower arm. Our real-world evaluation focuses on single-arm tabletop manipulation tasks.

\paragraph{Tasks.}
We evaluate eight real-world tabletop manipulation tasks spanning precision insertion, stacking, object placement, and multi-stage interaction: inserting a marker into a cup, stacking two blocks, placing two bottles into matching cups, standing a bottle upright, transferring a marker between cups, uncapping a marker, ranking three bottles by color, and arranging three bottles into a triangle. The first three tasks use 50 demonstration rollouts each, and the remaining five use 20 rollouts each. The appendix defines all tasks and shows additional qualitative rollouts. For every variant, we train one multi-task policy on the same demonstrations from all eight tasks. \model{} and \model{} w/o motion predict end-effector poses and are trained from the collected demonstrations only. The $\pi_{0.5}$ baseline predicts joint-angle actions and is fine-tuned with LoRA from its public base checkpoint. Section~\ref{sec:joint_angle} separately compares 2D-JA and \model{}-JA under the same joint-angle output parameterization.

\paragraph{Results.}
Results are summarized in Table~\ref{tab:real_world}. Across the eight tasks, \model{} achieves the highest mean success rate (62.5\%), outperforms \model{} without motion on every task, and exceeds $\pi_{0.5}$ on seven of eight tasks. Rank by color remains challenging. We hypothesize that its visually similar intermediate states to Place in triangle create cross-task interference, causing hesitation over the long horizon.
Qualitative rollouts show examples in which \model{} continues successfully after transient execution errors, partial occlusion, or small object-pose inaccuracies (Appendix Figure~\ref{fig:realworld_rollouts}).

\begin{table}[!ht]
\centering
\caption{\textbf{Real-world success rates on ALOHA2.}
Each entry reports the number of successful trials among 20 evaluations. \model{}-JA denotes the joint-angle action variant conditioned on 3D scene-motion features and intermediate end-effector action features; 2D-JA uses the same VLM, joint-angle denoiser, and training data but performs perception and motion reasoning in 2D; \model{} w/o motion removes the 3D Scene Trajectory Denoiser; \model{} is the full Cartesian-action model. \textsuperscript{\dag} denotes the three tasks with 50 demonstration rollouts each; the remaining five tasks use 20 demonstration rollouts.}
\begin{tabular}{@{}lccccc@{}}
\toprule
\textbf{Task} &
$\pi_{0.5}$ &
2D-JA &
\model{}-JA &
\model{} w/o motion &
\model{} \\
\midrule
Insert marker\textsuperscript{\dag}     & \underline{9/20} & 3/20 & \textbf{13/20} & 3/20 & 5/20 \\
Stack blocks\textsuperscript{\dag}      & 2/20 & 9/20 & \underline{17/20} & 15/20 & \textbf{19/20} \\
Place bottles\textsuperscript{\dag}     & 10/20 & \underline{16/20} & \underline{16/20} & 14/20 & \textbf{17/20} \\
Stand bottle      & 7/20 & 6/20 & \underline{13/20} & \underline{13/20} & \textbf{17/20} \\
Transfer marker   & 2/20 & 0/20 & \textbf{13/20} & 10/20 & \underline{12/20} \\
Uncap marker      & 15/20 & 12/20 & \underline{18/20} & 15/20 & \textbf{20/20} \\
Rank by color     & 0/20 & 0/20 & \underline{1/20} & 0/20 & \textbf{2/20} \\
Place in triangle & 2/20 & 0/20 & \underline{4/20} & 1/20 & \textbf{8/20} \\
\midrule
\textbf{Average}
                 & 29.4\%
                 & 28.8\%
                 & \underline{59.4\%}
                 & 44.4\%
                 & \textbf{62.5\%} \\
\bottomrule
\end{tabular}
\label{tab:real_world}
\end{table}

\subsection{3D Reasoning for Joint-Angle Robot Control}
\label{sec:joint_angle}
\model{} performs geometric reasoning over scene features, future motion, and end-effector actions in a shared 3D frame, but the executed action need not use a Cartesian parameterization. \model{}-JA adds a joint-angle denoiser conditioned on $f_{\ell}$, $f_v^{\mathrm{mot}}$, and the end-effector action features obtained after one velocity-field evaluation. These intermediate representations guide the prediction of executable joint-angle commands and allow us to compare 2D and 3D reasoning under a matched output action space. Joint-angle output avoids executing the predicted Cartesian trajectory directly.

\paragraph{Results.} As shown in Tables~\ref{tab:main_results} and~\ref{tab:real_world}, \model{}-JA and 2D-JA use the same joint-angle action parameterization and are trained on the same benchmark data. \model{}-JA outperforms 2D-JA on CALVIN (4.508 versus 4.195), the five-task RoboTwin2.0 subset (81.6\% versus 66.2\%), and the real-world evaluation (59.4\% versus 28.8\%). This matched comparison supports the benefit of the 3D scene-motion and end-effector action representations rather than joint-angle prediction alone. \model{}-JA also approaches the full Cartesian-action model on average and performs best on marker insertion (13/20, compared with 5/20 for \model{} and 9/20 for $\pi_{0.5}$). Thus, the benefit of \model{}'s geometric representation persists under both Cartesian and joint-angle output parameterizations.

\section{Limitations}
\label{sec:limitations}
 \model{} demonstrates strong performance across simulation and real-world manipulation benchmarks, but several limitations remain. 
First, it relies on accurate camera calibration; depth measurement errors or camera-extrinsic calibration errors can degrade the quality of geometric representations, particularly in real-world settings. Second, unlike recent large-scale VLA systems, \model{} is trained only on benchmark-scale robot demonstrations without large-scale robot pretraining or cross-embodiment data. Scaling to larger and more diverse datasets could further improve semantic generalization, robustness, and long-horizon reasoning. Finally, \model{} performs reasoning through latent geometric trajectory representations rather than explicit language-based reasoning. Combining geometric reasoning with language-conditioned planning or hierarchical task decomposition is a promising direction for long-horizon manipulation and failure recovery.

\section{Conclusion}
\label{sec:conclusion}

We presented \model{}, a 3D motion-aware vision-language-action framework that unifies semantic perception, latent future-motion reasoning, and action generation within a shared metric 3D coordinate frame. Our approach introduces a \emph{3D Scene Trajectory Denoiser} that predicts future scene motion directly in 3D space and integrates intermediate motion features into a flow-based 3D action denoiser through geometry-aware attention. Experiments across CALVIN, LIBERO, RoboTwin2.0, and real-world manipulation benchmarks show that maintaining geometric consistency between scene representation, future reasoning, and robot actions improves manipulation performance. These results motivate VLA architectures that couple semantic representations with physically grounded spatial and temporal reasoning.

%===============================================================================

\clearpage
% The acknowledgments are automatically included only in the final and preprint versions of the paper.
% \acknowledgments{If a paper is accepted, the final camera-ready version will (and probably should) include acknowledgments. All acknowledgments go at the end of the paper, including thanks to reviewers who gave useful comments, to colleagues who contributed to the ideas, and to funding agencies and corporate sponsors that provided financial support.}

%===============================================================================

% no \bibliographystyle is required, since the corl style is automatically used.
\bibliography{paper/camera_ready/reference}  % .bib

@misc{gkanatsios20253dflowmatchactorunified,
      title={3D FlowMatch Actor: Unified 3D Policy for Single- and Dual-Arm Manipulation}, 
      author={Nikolaos Gkanatsios and Jiahe Xu and Matthew Bronars and Arsalan Mousavian and Tsung-Wei Ke and Katerina Fragkiadaki},
      year={2025},
      eprint={2508.11002},
      archivePrefix={arXiv},
      primaryClass={cs.RO},
      url={https://arxiv.org/abs/2508.11002}, 
}

@misc{wang2026lamplearningvisionlanguageactionpolicies,
      title={LaMP: Learning Vision-Language-Action Policies with 3D Scene Flow as Latent Motion Prior}, 
      author={Xinkai Wang and Chenyi Wang and Yifu Xu and Mingzhe Ye and Fu-Cheng Zhang and Jialin Tian and Xinyu Zhan and Lifeng Zhu and Cewu Lu and Lixin Yang},
      year={2026},
      eprint={2603.25399},
      archivePrefix={arXiv},
      primaryClass={cs.CV},
      url={https://arxiv.org/abs/2603.25399}, 
}

@misc{xiao2023florence2advancingunifiedrepresentation,
      title={Florence-2: Advancing a Unified Representation for a Variety of Vision Tasks}, 
      author={Bin Xiao and Haiping Wu and Weijian Xu and Xiyang Dai and Houdong Hu and Yumao Lu and Michael Zeng and Ce Liu and Lu Yuan},
      year={2023},
      eprint={2311.06242},
      archivePrefix={arXiv},
      primaryClass={cs.CV},
      url={https://arxiv.org/abs/2311.06242}, 
}

@misc{xiao2025spatialtrackerv23dpointtracking,
      title={SpatialTrackerV2: 3D Point Tracking Made Easy}, 
      author={Yuxi Xiao and Jianyuan Wang and Nan Xue and Nikita Karaev and Yuri Makarov and Bingyi Kang and Xing Zhu and Hujun Bao and Yujun Shen and Xiaowei Zhou},
      year={2025},
      eprint={2507.12462},
      archivePrefix={arXiv},
      primaryClass={cs.CV},
      url={https://arxiv.org/abs/2507.12462}, 
}

@misc{li2025simplevlarlscalingvlatraining,
      title={SimpleVLA-RL: Scaling VLA Training via Reinforcement Learning}, 
      author={Haozhan Li and Yuxin Zuo and Jiale Yu and Yuhao Zhang and Zhaohui Yang and Kaiyan Zhang and Xuekai Zhu and Yuchen Zhang and Tianxing Chen and Ganqu Cui and Dehui Wang and Dingxiang Luo and Yuchen Fan and Youbang Sun and Jia Zeng and Jiangmiao Pang and Shanghang Zhang and Yu Wang and Yao Mu and Bowen Zhou and Ning Ding},
      year={2025},
      eprint={2509.09674},
      archivePrefix={arXiv},
      primaryClass={cs.RO},
      url={https://arxiv.org/abs/2509.09674}, 
}

@misc{brohan2023rt2visionlanguageactionmodelstransfer,
      title={RT-2: Vision-Language-Action Models Transfer Web Knowledge to Robotic Control}, 
      author={Anthony Brohan and Noah Brown and Justice Carbajal and Yevgen Chebotar and Xi Chen and Krzysztof Choromanski and Tianli Ding and Danny Driess and Avinava Dubey and Chelsea Finn and Pete Florence and Chuyuan Fu and Montse Gonzalez Arenas and Keerthana Gopalakrishnan and Kehang Han and Karol Hausman and Alexander Herzog and Jasmine Hsu and Brian Ichter and Alex Irpan and Nikhil Joshi and Ryan Julian and Dmitry Kalashnikov and Yuheng Kuang and Isabel Leal and Lisa Lee and Tsang-Wei Edward Lee and Sergey Levine and Yao Lu and Henryk Michalewski and Igor Mordatch and Karl Pertsch and Kanishka Rao and Krista Reymann and Michael Ryoo and Grecia Salazar and Pannag Sanketi and Pierre Sermanet and Jaspiar Singh and Anikait Singh and Radu Soricut and Huong Tran and Vincent Vanhoucke and Quan Vuong and Ayzaan Wahid and Stefan Welker and Paul Wohlhart and Jialin Wu and Fei Xia and Ted Xiao and Peng Xu and Sichun Xu and Tianhe Yu and Brianna Zitkovich},
      year={2023},
      eprint={2307.15818},
      archivePrefix={arXiv},
      primaryClass={cs.RO},
      url={https://arxiv.org/abs/2307.15818}, 
}

@misc{intelligence2025pi05visionlanguageactionmodelopenworld,
      title={$\pi_{0.5}$: a Vision-Language-Action Model with Open-World Generalization}, 
      author={Physical Intelligence and Kevin Black and Noah Brown and James Darpinian and Karan Dhabalia and Danny Driess and Adnan Esmail and Michael Equi and Chelsea Finn and Niccolo Fusai and Manuel Y. Galliker and Dibya Ghosh and Lachy Groom and Karol Hausman and Brian Ichter and Szymon Jakubczak and Tim Jones and Liyiming Ke and Devin LeBlanc and Sergey Levine and Adrian Li-Bell and Mohith Mothukuri and Suraj Nair and Karl Pertsch and Allen Z. Ren and Lucy Xiaoyang Shi and Laura Smith and Jost Tobias Springenberg and Kyle Stachowicz and James Tanner and Quan Vuong and Homer Walke and Anna Walling and Haohuan Wang and Lili Yu and Ury Zhilinsky},
      year={2025},
      eprint={2504.16054},
      archivePrefix={arXiv},
      primaryClass={cs.LG},
      url={https://arxiv.org/abs/2504.16054}, 
}

@misc{black2026pi0visionlanguageactionflowmodel,
      title={$\pi_0$: A Vision-Language-Action Flow Model for General Robot Control}, 
      author={Kevin Black and Noah Brown and Danny Driess and Adnan Esmail and Michael Equi and Chelsea Finn and Niccolo Fusai and Lachy Groom and Karol Hausman and Brian Ichter and Szymon Jakubczak and Tim Jones and Liyiming Ke and Sergey Levine and Adrian Li-Bell and Mohith Mothukuri and Suraj Nair and Karl Pertsch and Lucy Xiaoyang Shi and James Tanner and Quan Vuong and Anna Walling and Haohuan Wang and Ury Zhilinsky},
      year={2026},
      eprint={2410.24164},
      archivePrefix={arXiv},
      primaryClass={cs.LG},
      url={https://arxiv.org/abs/2410.24164}, 
}

@misc{beyer2024paligemmaversatile3bvlm,
      title={PaliGemma: A versatile 3B VLM for transfer}, 
      author={Lucas Beyer and Andreas Steiner and André Susano Pinto and Alexander Kolesnikov and Xiao Wang and Daniel Salz and Maxim Neumann and Ibrahim Alabdulmohsin and Michael Tschannen and Emanuele Bugliarello and Thomas Unterthiner and Daniel Keysers and Skanda Koppula and Fangyu Liu and Adam Grycner and Alexey Gritsenko and Neil Houlsby and Manoj Kumar and Keran Rong and Julian Eisenschlos and Rishabh Kabra and Matthias Bauer and Matko Bošnjak and Xi Chen and Matthias Minderer and Paul Voigtlaender and Ioana Bica and Ivana Balazevic and Joan Puigcerver and Pinelopi Papalampidi and Olivier Henaff and Xi Xiong and Radu Soricut and Jeremiah Harmsen and Xiaohua Zhai},
      year={2024},
      eprint={2407.07726},
      archivePrefix={arXiv},
      primaryClass={cs.CV},
      url={https://arxiv.org/abs/2407.07726}, 
}

@misc{qwen2025qwen25technicalreport,
      title={Qwen2.5 Technical Report}, 
      author={Qwen and : and An Yang and Baosong Yang and Beichen Zhang and Binyuan Hui and Bo Zheng and Bowen Yu and Chengyuan Li and Dayiheng Liu and Fei Huang and Haoran Wei and Huan Lin and Jian Yang and Jianhong Tu and Jianwei Zhang and Jianxin Yang and Jiaxi Yang and Jingren Zhou and Junyang Lin and Kai Dang and Keming Lu and Keqin Bao and Kexin Yang and Le Yu and Mei Li and Mingfeng Xue and Pei Zhang and Qin Zhu and Rui Men and Runji Lin and Tianhao Li and Tianyi Tang and Tingyu Xia and Xingzhang Ren and Xuancheng Ren and Yang Fan and Yang Su and Yichang Zhang and Yu Wan and Yuqiong Liu and Zeyu Cui and Zhenru Zhang and Zihan Qiu},
      year={2025},
      eprint={2412.15115},
      archivePrefix={arXiv},
      primaryClass={cs.CL},
      url={https://arxiv.org/abs/2412.15115}, 
}

@article{gervet2023act3d,
  title={Act3D: 3D Feature Field Transformers for Multi-Task Robotic Manipulation},
  author={Gervet, Theophile and Xian, Zhou and Gkanatsios, Nikolaos and Fragkiadaki, Katerina},
  journal={CoRL},
  year={2023}
}

@inproceedings{shridhar2023perceiver,
  title={Perceiver-actor: A multi-task transformer for robotic manipulation},
  author={Shridhar, Mohit and Manuelli, Lucas and Fox, Dieter},
  booktitle={Conference on Robot Learning},
  pages={785--799},
  year={2023},
  organization={PMLR}
}

@article{ke20243d,
  title={3d diffuser actor: Policy diffusion with 3d scene representations},
  author={Ke, Tsung-Wei and Gkanatsios, Nikolaos and Fragkiadaki, Katerina},
  journal={arXiv preprint arXiv:2402.10885},
  year={2024}
}

@article{goyal2023rvt,
  title={RVT: Robotic View Transformer for 3D Object Manipulation},
  author={Goyal, Ankit and Xu, Jie and Guo, Yijie and Blukis, Valts and Chao, Yu-Wei and Fox, Dieter},
  journal={arXiv preprint arXiv:2306.14896},
  year={2023}
}

@misc{reuss2025flowerdemocratizinggeneralistrobot,
      title={FLOWER: Democratizing Generalist Robot Policies with Efficient Vision-Language-Action Flow Policies}, 
      author={Moritz Reuss and Hongyi Zhou and Marcel Rühle and Ömer Erdinç Yağmurlu and Fabian Otto and Rudolf Lioutikov},
      year={2025},
      eprint={2509.04996},
      archivePrefix={arXiv},
      primaryClass={cs.RO},
      url={https://arxiv.org/abs/2509.04996}, 
}

@misc{zheng2025xvlasoftpromptedtransformerscalable,
      title={X-VLA: Soft-Prompted Transformer as Scalable Cross-Embodiment Vision-Language-Action Model}, 
      author={Jinliang Zheng and Jianxiong Li and Zhihao Wang and Dongxiu Liu and Xirui Kang and Yuchun Feng and Yinan Zheng and Jiayin Zou and Yilun Chen and Jia Zeng and Ya-Qin Zhang and Jiangmiao Pang and Jingjing Liu and Tai Wang and Xianyuan Zhan},
      year={2025},
      eprint={2510.10274},
      archivePrefix={arXiv},
      primaryClass={cs.RO},
      url={https://arxiv.org/abs/2510.10274}, 
}

@misc{zhen20243dvla3dvisionlanguageactiongenerative,
      title={3D-VLA: A 3D Vision-Language-Action Generative World Model}, 
      author={Haoyu Zhen and Xiaowen Qiu and Peihao Chen and Jincheng Yang and Xin Yan and Yilun Du and Yining Hong and Chuang Gan},
      year={2024},
      eprint={2403.09631},
      archivePrefix={arXiv},
      primaryClass={cs.CV},
      url={https://arxiv.org/abs/2403.09631}, 
}

@misc{qu2025spatialvlaexploringspatialrepresentations,
      title={SpatialVLA: Exploring Spatial Representations for Visual-Language-Action Model}, 
      author={Delin Qu and Haoming Song and Qizhi Chen and Yuanqi Yao and Xinyi Ye and Yan Ding and Zhigang Wang and JiaYuan Gu and Bin Zhao and Dong Wang and Xuelong Li},
      year={2025},
      eprint={2501.15830},
      archivePrefix={arXiv},
      primaryClass={cs.RO},
      url={https://arxiv.org/abs/2501.15830}, 
}

@misc{zhang20254dvlaspatiotemporalvisionlanguageactionpretraining,
      title={4D-VLA: Spatiotemporal Vision-Language-Action Pretraining with Cross-Scene Calibration}, 
      author={Jiahui Zhang and Yurui Chen and Yueming Xu and Ze Huang and Yanpeng Zhou and Yu-Jie Yuan and Xinyue Cai and Guowei Huang and Xingyue Quan and Hang Xu and Li Zhang},
      year={2025},
      eprint={2506.22242},
      archivePrefix={arXiv},
      primaryClass={cs.CV},
      url={https://arxiv.org/abs/2506.22242}, 
}

@misc{jia2024lift3dfoundationpolicylifting,
      title={Lift3D Foundation Policy: Lifting 2D Large-Scale Pretrained Models for Robust 3D Robotic Manipulation}, 
      author={Yueru Jia and Jiaming Liu and Sixiang Chen and Chenyang Gu and Zhilue Wang and Longzan Luo and Lily Lee and Pengwei Wang and Zhongyuan Wang and Renrui Zhang and Shanghang Zhang},
      year={2024},
      eprint={2411.18623},
      archivePrefix={arXiv},
      primaryClass={cs.CV},
      url={https://arxiv.org/abs/2411.18623}, 
}

@misc{li2025pointvlainjecting3dworld,
      title={PointVLA: Injecting the 3D World into Vision-Language-Action Models}, 
      author={Chengmeng Li and Junjie Wen and Yan Peng and Yaxin Peng and Feifei Feng and Yichen Zhu},
      year={2025},
      eprint={2503.07511},
      archivePrefix={arXiv},
      primaryClass={cs.RO},
      url={https://arxiv.org/abs/2503.07511}, 
}

@misc{cen2025worldvlaautoregressiveactionworld,
      title={WorldVLA: Towards Autoregressive Action World Model}, 
      author={Jun Cen and Chaohui Yu and Hangjie Yuan and Yuming Jiang and Siteng Huang and Jiayan Guo and Xin Li and Yibing Song and Hao Luo and Fan Wang and Deli Zhao and Hao Chen},
      year={2025},
      eprint={2506.21539},
      archivePrefix={arXiv},
      primaryClass={cs.RO},
      url={https://arxiv.org/abs/2506.21539}, 
}

@misc{zhong2025flowvlavisualchainthoughtbased,
      title={FlowVLA: Visual Chain of Thought-based Motion Reasoning for Vision-Language-Action Models}, 
      author={Zhide Zhong and Haodong Yan and Junfeng Li and Xiangchen Liu and Xin Gong and Tianran Zhang and Wenxuan Song and Jiayi Chen and Xinhu Zheng and Hesheng Wang and Haoang Li},
      year={2025},
      eprint={2508.18269},
      archivePrefix={arXiv},
      primaryClass={cs.RO},
      url={https://arxiv.org/abs/2508.18269}, 
}

@misc{zheng2025tracevlavisualtraceprompting,
      title={TraceVLA: Visual Trace Prompting Enhances Spatial-Temporal Awareness for Generalist Robotic Policies}, 
      author={Ruijie Zheng and Yongyuan Liang and Shuaiyi Huang and Jianfeng Gao and Hal Daumé III and Andrey Kolobov and Furong Huang and Jianwei Yang},
      year={2025},
      eprint={2412.10345},
      archivePrefix={arXiv},
      primaryClass={cs.RO},
      url={https://arxiv.org/abs/2412.10345}, 
}

@misc{gu2023rttrajectoryrobotictaskgeneralization,
      title={RT-Trajectory: Robotic Task Generalization via Hindsight Trajectory Sketches}, 
      author={Jiayuan Gu and Sean Kirmani and Paul Wohlhart and Yao Lu and Montserrat Gonzalez Arenas and Kanishka Rao and Wenhao Yu and Chuyuan Fu and Keerthana Gopalakrishnan and Zhuo Xu and Priya Sundaresan and Peng Xu and Hao Su and Karol Hausman and Chelsea Finn and Quan Vuong and Ted Xiao},
      year={2023},
      eprint={2311.01977},
      archivePrefix={arXiv},
      primaryClass={cs.RO},
      url={https://arxiv.org/abs/2311.01977}, 
}

@article{mees2022calvin,
    author = {Oier Mees and Lukas Hermann and Erick Rosete-Beas and Wolfram Burgard},
    title = {CALVIN: A Benchmark for Language-Conditioned Policy Learning for Long-Horizon Robot Manipulation Tasks},
    journal={IEEE Robotics and Automation Letters (RA-L)},
    volume={7},
    number={3},
    pages={7327-7334},
    year={2022}
}

@article{liu2024libero,
  title={Libero: Benchmarking knowledge transfer for lifelong robot learning},
  author={Liu, Bo and Zhu, Yifeng and Gao, Chongkai and Feng, Yihao and Liu, Qiang and Zhu, Yuke and Stone, Peter},
  journal={Advances in Neural Information Processing Systems},
  volume={36},
  year={2024}
}

@misc{chen2025robotwin20scalabledata,
      title={RoboTwin 2.0: A Scalable Data Generator and Benchmark with Strong Domain Randomization for Robust Bimanual Robotic Manipulation}, 
      author={Tianxing Chen and Zanxin Chen and Baijun Chen and Zijian Cai and Yibin Liu and Zixuan Li and Qiwei Liang and Xianliang Lin and Yiheng Ge and Zhenyu Gu and Weiliang Deng and Yubin Guo and Tian Nian and Xuanbing Xie and Qiangyu Chen and Kailun Su and Tianling Xu and Guodong Liu and Mengkang Hu and Huan-ang Gao and Kaixuan Wang and Zhixuan Liang and Yusen Qin and Xiaokang Yang and Ping Luo and Yao Mu},
      year={2025},
      eprint={2506.18088},
      archivePrefix={arXiv},
      primaryClass={cs.RO},
      url={https://arxiv.org/abs/2506.18088}, 
}

@article{chisari2024learning,
  title={Learning robotic manipulation policies from point clouds with conditional flow matching},
  author={Chisari, Eugenio and Heppert, Nick and Argus, Max and Welschehold, Tim and Brox, Thomas and Valada, Abhinav},
  journal={arXiv preprint arXiv:2409.07343},
  year={2024}
}

@article{Liu2024VoxActBVA,
  title={VoxAct-B: Voxel-Based Acting and Stabilizing Policy for Bimanual Manipulation},
  author={I-Chun Arthur Liu and Sicheng He and Daniel Seita and Gaurav Sukhatme},
  journal={CoRL},
  year={2024}
}

@article{grotz2024peract2,
  title={PerAct2: A Perceiver Actor Framework for Bimanual Manipulation Tasks},
  author={Grotz, Markus and Shridhar, Mohit and Asfour, Tamim and Fox, Dieter},
  journal={arXiv preprint arXiv:2407.00278},
  year={2024}
}

@inproceedings{xian2023chaineddiffuser,
  title={Chaineddiffuser: Unifying trajectory diffusion and keypose prediction for robotic manipulation},
  author={Xian, Zhou and Gkanatsios, Nikolaos and Gervet, Theophile and Ke, Tsung-Wei and Fragkiadaki, Katerina},
  booktitle={Conference on Robot Learning},
  pages={2323--2339},
  year={2023},
  organization={PMLR}
}

@misc{li2025bridgevlainputoutputalignmentefficient,
      title={BridgeVLA: Input-Output Alignment for Efficient 3D Manipulation Learning with Vision-Language Models}, 
      author={Peiyan Li and Yixiang Chen and Hongtao Wu and Xiao Ma and Xiangnan Wu and Yan Huang and Liang Wang and Tao Kong and Tieniu Tan},
      year={2025},
      eprint={2506.07961},
      archivePrefix={arXiv},
      primaryClass={cs.RO},
      url={https://arxiv.org/abs/2506.07961}, 
}

@misc{lin2025evo0visionlanguageactionmodelimplicit,
      title={Evo-0: Vision-Language-Action Model with Implicit Spatial Understanding}, 
      author={Tao Lin and Gen Li and Yilei Zhong and Yanwen Zou and Yuxin Du and Jiting Liu and Encheng Gu and Bo Zhao},
      year={2025},
      eprint={2507.00416},
      archivePrefix={arXiv},
      primaryClass={cs.RO},
      url={https://arxiv.org/abs/2507.00416}, 
}

@misc{li2026consisvla4dadvancingspatiotemporalconsistency,
      title={ConsisVLA-4D: Advancing Spatiotemporal Consistency in Efficient 3D-Perception and 4D-Reasoning for Robotic Manipulation}, 
      author={Wei Li and Jizhihui Liu and Li Yixing and Junwen Tong and Rui Shao and Liqiang Nie},
      year={2026},
      eprint={2605.05126},
      archivePrefix={arXiv},
      primaryClass={cs.RO},
      url={https://arxiv.org/abs/2605.05126}, 
}

@misc{fan2026any3dvlaenhancingvlarobustness,
      title={Any3D-VLA: Enhancing VLA Robustness via Diverse Point Clouds}, 
      author={Xianzhe Fan and Shengliang Deng and Xiaoyang Wu and Yuxiang Lu and Zhuoling Li and Mi Yan and Yujia Zhang and Zhizheng Zhang and He Wang and Hengshuang Zhao},
      year={2026},
      eprint={2602.00807},
      archivePrefix={arXiv},
      primaryClass={cs.CV},
      url={https://arxiv.org/abs/2602.00807}, 
}

@misc{deng2025stereovlaenhancingvisionlanguageactionmodels,
      title={StereoVLA: Enhancing Vision-Language-Action Models with Stereo Vision}, 
      author={Shengliang Deng and Mi Yan and Yixin Zheng and Jiayi Su and Wenhao Zhang and Xiaoguang Zhao and Heming Cui and Zhizheng Zhang and He Wang},
      year={2025},
      eprint={2512.21970},
      archivePrefix={arXiv},
      primaryClass={cs.RO},
      url={https://arxiv.org/abs/2512.21970}, 
}

@misc{li2025unifiedvideoactionmodel,
      title={Unified Video Action Model}, 
      author={Shuang Li and Yihuai Gao and Dorsa Sadigh and Shuran Song},
      year={2025},
      eprint={2503.00200},
      archivePrefix={arXiv},
      primaryClass={cs.RO},
      url={https://arxiv.org/abs/2503.00200}, 
}

@misc{du2023learninguniversalpoliciestextguided,
      title={Learning Universal Policies via Text-Guided Video Generation}, 
      author={Yilun Du and Mengjiao Yang and Bo Dai and Hanjun Dai and Ofir Nachum and Joshua B. Tenenbaum and Dale Schuurmans and Pieter Abbeel},
      year={2023},
      eprint={2302.00111},
      archivePrefix={arXiv},
      primaryClass={cs.AI},
      url={https://arxiv.org/abs/2302.00111}, 
}

@misc{bi2025motusunifiedlatentaction,
      title={Motus: A Unified Latent Action World Model}, 
      author={Hongzhe Bi and Hengkai Tan and Shenghao Xie and Zeyuan Wang and Shuhe Huang and Haitian Liu and Ruowen Zhao and Yao Feng and Chendong Xiang and Yinze Rong and Hongyan Zhao and Hanyu Liu and Zhizhong Su and Lei Ma and Hang Su and Jun Zhu},
      year={2025},
      eprint={2512.13030},
      archivePrefix={arXiv},
      primaryClass={cs.CV},
      url={https://arxiv.org/abs/2512.13030}, 
}

@misc{gigabrainteam2026gigabrain05mvlalearnsworld,
      title={GigaBrain-0.5M*: a VLA That Learns From World Model-Based Reinforcement Learning}, 
      author={GigaBrain Team and Boyuan Wang and Bohan Li and Chaojun Ni and Guan Huang and Guosheng Zhao and Hao Li and Jie Li and Jindi Lv and Jingyu Liu and Lv Feng and Mingming Yu and Peng Li and Qiuping Deng and Tianze Liu and Xinyu Zhou and Xinze Chen and Xiaofeng Wang and Yang Wang and Yifan Li and Yifei Nie and Yilong Li and Yukun Zhou and Yun Ye and Zhichao Liu and Zheng Zhu},
      year={2026},
      eprint={2602.12099},
      archivePrefix={arXiv},
      primaryClass={cs.CV},
      url={https://arxiv.org/abs/2602.12099}, 
}

@misc{lin2026roboflow4dlightweightflowworld,
      title={RoboFlow4D: A Lightweight Flow World Model Toward Real-Time Flow-Guided Robotic Manipulation}, 
      author={Sixu Lin and Junliang Chen and Huaiyuan Xu and Zhuohao Li and Guangming Wang and Yixiong Jing and Sheng Xu and Runyi Zhao and Brian Sheil and Lap-Pui Chau and Guiliang Liu},
      year={2026},
      eprint={2605.17522},
      archivePrefix={arXiv},
      primaryClass={cs.RO},
      url={https://arxiv.org/abs/2605.17522}, 
}

@misc{zhi20253dflowactionlearningcrossembodimentmanipulation,
      title={3DFlowAction: Learning Cross-Embodiment Manipulation from 3D Flow World Model}, 
      author={Hongyan Zhi and Peihao Chen and Siyuan Zhou and Yubo Dong and Quanxi Wu and Lei Han and Mingkui Tan},
      year={2025},
      eprint={2506.06199},
      archivePrefix={arXiv},
      primaryClass={cs.RO},
      url={https://arxiv.org/abs/2506.06199}, 
}

@misc{lee2025tracegenworldmodeling3d,
      title={TraceGen: World Modeling in 3D Trace Space Enables Learning from Cross-Embodiment Videos}, 
      author={Seungjae Lee and Yoonkyo Jung and Inkook Chun and Yao-Chih Lee and Zikui Cai and Hongjia Huang and Aayush Talreja and Tan Dat Dao and Yongyuan Liang and Jia-Bin Huang and Furong Huang},
      year={2025},
      eprint={2511.21690},
      archivePrefix={arXiv},
      primaryClass={cs.RO},
      url={https://arxiv.org/abs/2511.21690}, 
}

@misc{noh20253dflowdiffusionpolicy,
      title={3D Flow Diffusion Policy: Visuomotor Policy Learning via Generating Flow in 3D Space}, 
      author={Sangjun Noh and Dongwoo Nam and Kangmin Kim and Geonhyup Lee and Yeonguk Yu and Raeyoung Kang and Kyoobin Lee},
      year={2025},
      eprint={2509.18676},
      archivePrefix={arXiv},
      primaryClass={cs.RO},
      url={https://arxiv.org/abs/2509.18676}, 
}

@misc{xu2024flowcrossdomainmanipulationinterface,
      title={Flow as the Cross-Domain Manipulation Interface}, 
      author={Mengda Xu and Zhenjia Xu and Yinghao Xu and Cheng Chi and Gordon Wetzstein and Manuela Veloso and Shuran Song},
      year={2024},
      eprint={2407.15208},
      archivePrefix={arXiv},
      primaryClass={cs.RO},
      url={https://arxiv.org/abs/2407.15208}, 
}

@misc{lee2025molmoactactionreasoningmodels,
      title={MolmoAct: Action Reasoning Models that can Reason in Space}, 
      author={Jason Lee and Jiafei Duan and Haoquan Fang and Yuquan Deng and Shuo Liu and Boyang Li and Bohan Fang and Jieyu Zhang and Yi Ru Wang and Sangho Lee and Winson Han and Wilbert Pumacay and Angelica Wu and Rose Hendrix and Karen Farley and Eli VanderBilt and Ali Farhadi and Dieter Fox and Ranjay Krishna},
      year={2025},
      eprint={2508.07917},
      archivePrefix={arXiv},
      primaryClass={cs.RO},
      url={https://arxiv.org/abs/2508.07917}, 
}

@misc{li2025qdepthvlaquantizeddepthprediction,
      title={QDepth-VLA: Quantized Depth Prediction as Auxiliary Supervision for Vision-Language-Action Models}, 
      author={Yixuan Li and Yuhui Chen and Mingcai Zhou and Haoran Li and Zhengtao Zhang and Dongbin Zhao},
      year={2025},
      eprint={2510.14836},
      archivePrefix={arXiv},
      primaryClass={cs.CV},
      url={https://arxiv.org/abs/2510.14836}, 
}

@misc{li2025hamsterhierarchicalactionmodels,
      title={HAMSTER: Hierarchical Action Models For Open-World Robot Manipulation}, 
      author={Yi Li and Yuquan Deng and Jesse Zhang and Joel Jang and Marius Memmel and Raymond Yu and Caelan Reed Garrett and Fabio Ramos and Dieter Fox and Anqi Li and Abhishek Gupta and Ankit Goyal},
      year={2025},
      eprint={2502.05485},
      archivePrefix={arXiv},
      primaryClass={cs.RO},
      url={https://arxiv.org/abs/2502.05485}, 
}

@misc{bharadhwaj2024track2actpredictingpointtracks,
      title={Track2Act: Predicting Point Tracks from Internet Videos enables Generalizable Robot Manipulation}, 
      author={Homanga Bharadhwaj and Roozbeh Mottaghi and Abhinav Gupta and Shubham Tulsiani},
      year={2024},
      eprint={2405.01527},
      archivePrefix={arXiv},
      primaryClass={cs.RO},
      url={https://arxiv.org/abs/2405.01527}, 
}

@misc{wen2024anypointtrajectorymodelingpolicy,
      title={Any-point Trajectory Modeling for Policy Learning}, 
      author={Chuan Wen and Xingyu Lin and John So and Kai Chen and Qi Dou and Yang Gao and Pieter Abbeel},
      year={2024},
      eprint={2401.00025},
      archivePrefix={arXiv},
      primaryClass={cs.RO},
      url={https://arxiv.org/abs/2401.00025}, 
}

@misc{zawalski2025roboticcontrolembodiedchainofthought,
      title={Robotic Control via Embodied Chain-of-Thought Reasoning}, 
      author={Michał Zawalski and William Chen and Karl Pertsch and Oier Mees and Chelsea Finn and Sergey Levine},
      year={2025},
      eprint={2407.08693},
      archivePrefix={arXiv},
      primaryClass={cs.RO},
      url={https://arxiv.org/abs/2407.08693}, 
}

@misc{kim2025finetuningvisionlanguageactionmodelsoptimizing,
      title={Fine-Tuning Vision-Language-Action Models: Optimizing Speed and Success}, 
      author={Moo Jin Kim and Chelsea Finn and Percy Liang},
      year={2025},
      eprint={2502.19645},
      archivePrefix={arXiv},
      primaryClass={cs.RO},
      url={https://arxiv.org/abs/2502.19645}, 
}

@misc{lin2026universalposepretraininggeneralizable,
      title={Universal Pose Pretraining for Generalizable Vision-Language-Action Policies}, 
      author={Haitao Lin and Hanyang Yu and Jingshun Huang and He Zhang and Yonggen Ling and Ping Tan and Xiangyang Xue and Yanwei Fu},
      year={2026},
      eprint={2602.19710},
      archivePrefix={arXiv},
      primaryClass={cs.CV},
      url={https://arxiv.org/abs/2602.19710}, 
}

@misc{pai2025mimicvideovideoactionmodelsgeneralizable,
      title={mimic-video: Video-Action Models for Generalizable Robot Control Beyond VLAs}, 
      author={Jonas Pai and Liam Achenbach and Victoriano Montesinos and Benedek Forrai and Oier Mees and Elvis Nava},
      year={2025},
      eprint={2512.15692},
      archivePrefix={arXiv},
      primaryClass={cs.RO},
      url={https://arxiv.org/abs/2512.15692}, 
}

@misc{zhao2025cotvlavisualchainofthoughtreasoning,
      title={CoT-VLA: Visual Chain-of-Thought Reasoning for Vision-Language-Action Models}, 
      author={Qingqing Zhao and Yao Lu and Moo Jin Kim and Zipeng Fu and Zhuoyang Zhang and Yecheng Wu and Zhaoshuo Li and Qianli Ma and Song Han and Chelsea Finn and Ankur Handa and Ming-Yu Liu and Donglai Xiang and Gordon Wetzstein and Tsung-Yi Lin},
      year={2025},
      eprint={2503.22020},
      archivePrefix={arXiv},
      primaryClass={cs.CV},
      url={https://arxiv.org/abs/2503.22020}, 
}

@inproceedings{sun2026geovla,
  title     = {GeoVLA: Empowering 3D Representations in Vision-Language-Action Models},
  author    = {Sun, Lin and Xie, Bin and Liu, Yingfei and Shi, Hao and Wang, Tiancai and Cao, Jiale},
  booktitle = {Proceedings of the IEEE/RSJ International Conference on Intelligent Robots and Systems (IROS)},
  year      = {2026}
}

@misc{yang2025fp33dfoundationpolicy,
      title={FP3: A 3D Foundation Policy for Robotic Manipulation}, 
      author={Rujia Yang and Geng Chen and Chuan Wen and Yang Gao},
      year={2025},
      archivePrefix={arXiv},
      primaryClass={cs.RO}
}

@inproceedings{hsu2025spot,
  author    = {Hsu, Cheng-Chun and Wen, Bowen and Xu, Jie and Narang, Yashraj and Wang, Xiaolong and Zhu, Yuke and Biswas, Joydeep and Birchfield, Stan},
  title     = {SPOT: SE(3) Pose Trajectory Diffusion for Object-Centric Manipulation},
  booktitle = {IEEE International Conference on Robotics and Automation (ICRA)},
  year      = {2025},
}

@misc{vosylius2024renderdiffusealigningimage,
      title={Render and Diffuse: Aligning Image and Action Spaces for Diffusion-based Behaviour Cloning}, 
      author={Vitalis Vosylius and Younggyo Seo and Jafar Uruç and Stephen James},
      year={2024},
      archivePrefix={arXiv},
      primaryClass={cs.RO}
}

\clearpage
\appendix
\section*{Appendix}
\paragraph{Model configurations and compute.}
Table~\ref{tab:model_config} reports the model configuration used for each benchmark. All variants follow the same architectural design, while hidden dimensions, layer counts, action horizons, batch sizes, and optimization schedules are adapted to each benchmark. The second-stage joint training requires approximately 15 hours on four 48 GB NVIDIA L40S GPUs for CALVIN, 25 hours on eight 40 GB NVIDIA A100 GPUs for LIBERO, and 48 hours on eight 40 GB NVIDIA A100 GPUs for the five-task RoboTwin2.0 ablation setting.

\paragraph{Lifting 2D VLM features into 3D.}
We assign each patch token a depth value by bilinearly interpolating the four neighboring depth samples at the patch center. Camera intrinsics and extrinsics then map the patch center to a 3D position in the robot base frame.

\paragraph{3D scene trajectory encoder.}
The encoder binds VLM-derived semantics to local 3D geometry through a point-wise design. Each image query $\mathbf{u}_i=(u_i,v_i)$ is lifted to a base-frame point $\mathbf{q}_i$. We use centered coordinates $\tilde{\mathbf{q}}_i$ inside the trajectory encoder to remove global translation, while retaining the uncentered base-frame positions $\mathbf{q}_i$ for downstream spatial encoding:
\begin{equation}
\bar{\mathbf{u}}_i=[u_i,v_i,1]^\top,\qquad
\mathbf{q}_i=R_{c\rightarrow b}\bigl(D(\mathbf{u}_i)K^{-1}\bar{\mathbf{u}}_i\bigr)+\mathbf{t}_{c\rightarrow b},\qquad
\tilde{\mathbf{q}}_i=\mathbf{q}_i-\frac{1}{N}\sum_{j=1}^N\mathbf{q}_j.
\end{equation}
Here $K$ is the camera-intrinsic matrix, $D(\mathbf{u}_i)$ is the interpolated depth, and $R_{c\rightarrow b}$ and $\mathbf{t}_{c\rightarrow b}$ are the camera-to-base rotation and translation.
A frozen Florence-2 VLM encodes the image into a patch feature map
$V\in\mathbb{R}^{H_v\times W_v\times d_v}$ and the instruction into language
tokens $E\in\mathbb{R}^{L\times d_t}$. For each query we bilinearly sample $V$
to obtain a visual descriptor $x_i$ and form an initial point token that fuses
geometry, image location, and semantics:
\begin{equation}
h_i^{(0)} \;=\; \mathrm{MLP}\!\big([\,\tilde{\mathbf{q}}_i;\; \mathbf{u}_i;\; x_i\,]\big)
\;\in\; \mathbb{R}^{D}.
\end{equation}
We build a $k$-nearest-neighbor graph $\mathcal{N}_k(i)$ once over
$\{\tilde{\mathbf{q}}_i\}$ and refine tokens through $L_{\mathrm{enc}}$ local
geometry-aware attention layers, where each query attends to its neighbors with
attention logits augmented by a learned relative-position bias
$b(\tilde{\mathbf{q}}_j-\tilde{\mathbf{q}}_i)$:
\begin{equation}
h_i^{(\ell+1)}
\;=\; h_i^{(\ell)} + \!\!\sum_{j\in\mathcal{N}_k(i)}\!\!
\alpha_{ij}^{(\ell)}\, W_v^{(\ell)} h_j^{(\ell)},
\qquad
\alpha_{ij}^{(\ell)} \propto
\exp\!\Bigg(
\frac{\bigl(W_q^{(\ell)} h_i^{(\ell)}\bigr)^{\!\top}\!\bigl(W_k^{(\ell)} h_j^{(\ell)}\bigr)}
     {\sqrt{d_h}}
+ b(\tilde{\mathbf{q}}_j - \tilde{\mathbf{q}}_i)
\Bigg).
\end{equation}
Here $d_h$ is the per-head key dimension. The encoder outputs per-point context features $\{c_i\}_{i=1}^{N}$ that couple task-relevant semantics with local 3D structure. The decoder also receives the VLM's pooled front-view visual tokens $V_g$ and projected language tokens $E$ as global multimodal conditioning, $\mathcal{C}=(\{c_i\},V_g,E)$.

\paragraph{3D scene trajectory decoder.}
The decoder predicts $\mathcal{F}$ by conditional flow matching.
% in the normalized displacement space. 
Let $F\in\mathbb{R}^{N\times H\times 3}$ denote the normalized pseudo-ground-truth displacement tensor extracted by SpatialTrackerV2. Displacements are mapped into the normalized training coordinates before flow matching and mapped back with the inverse transformation at inference.
We sample Gaussian noise $F_0\sim\mathcal{N}(0,I)$ and a flow time $\tau\sim\mathcal{U}(0,1)$, and define the linear interpolant and the velocity target
\begin{equation}
F_\tau \;=\; (1-\tau)\,F_0 + \tau\,F,
\qquad
u^{\star} \;=\; F - F_0.
\end{equation}
A velocity network $u_\theta(F_\tau, \tau, \mathcal{C})$ is trained to regress $u^{\star}$, where $\mathcal{C} = (\{c_i\}, V_g, E)$ denotes the multimodal context.

The decoder input at point $i$ and step $t$ fuses the noisy displacement, the
per-point context, a learned future-time-step embedding $e_t$ and a sinusoidal flow-time
embedding $\phi(\tau)$,
\begin{equation}
Z_{i,t}^{(0)} \;=\; W_F\, F_{\tau, i, t} \;+\; c_i \;+\; e_t \;+\; \phi(\tau).
\end{equation}

Each decoder block factorizes attention along the spatial and temporal axes of the trajectory field.
Spatial attention operates over 3D anchors at each future step.
% natural axes of the motion tensor and then integrates multimodal conditioning.
Rather than re-using the encoder's kNN graph, we adopt a serialized window
attention along a 3D space-filling curve, which scales to the full $N\!\times\!H$
token set with hardware-friendly fixed-size windows. 
Concretely, for each future step $t$ we normalize the anchors to the unit cube,
$\hat{q}_i = (\tilde{q}_i - \min_j \tilde{q}_j)/(\max_j \tilde{q}_j - \min_j \tilde{q}_j)$,
quantize them to an integer grid at bit-depth $d_q$, and assign each anchor a
3D Morton (Z-order) code $\mathrm{m}(\hat{q}_i)\in\mathbb{N}$ that yields a
permutation $\pi$ sorting the $N$ points along the curve. Self-attention is
then applied within non-overlapping windows of $P$ consecutive points in the
order $\pi$, and the result is restored to the original point order by the
inverse permutation $\pi^{-1}$:
\begin{equation}
\mathrm{SerSpatial}(Z)_{\pi(i), t}
\;=\; \mathrm{Attn}\!\big(Z_{\pi(i), t} \,;\; \{ Z_{\pi(j),t} : j \in \mathcal{W}_P(i) \}\big),
\qquad
\mathcal{W}_P(i) = \big\{j : \lfloor j/P \rfloor = \lfloor i/P \rfloor\big\}.
\end{equation}
To enlarge the effective receptive field while keeping per-window cost
$\mathcal{O}(P^2)$, we alternate between canonical $(x,y,z)$ and axis-permuted $(y,x,z)$ Morton orderings and apply a Swin-style half-window shift on
alternating layers; a learned linear projection of the normalized coordinates
is added to each token before the query, key, and value projections so that serialized attention
remains explicitly geometry-aware.

For every point $i$, a temporal self-attention $\mathrm{Temporal}(\cdot)$ couples
its $H$ future steps. The resulting tokens then attend to the conditioning
sequence $[V_g;\,E]$ through a cross-attention block
$\mathrm{Cross}(\cdot, [V_g; E])$, followed by a position-wise feed-forward
$\mathrm{FFN}(\cdot)$. With pre-LayerNorm and residual connections, one block
computes
\begin{equation}
Z \;\xrightarrow{\;\mathrm{SerSpatial}\;}\;
Z \;\xrightarrow{\;\mathrm{Temporal}\;}\;
Z \;\xrightarrow{\;\mathrm{Cross}([V_g;E])\;}\;
Z \;\xrightarrow{\;\mathrm{FFN}\;}\; Z.
\end{equation}
A linear head reads off the predicted velocity
$u_\theta(F_\tau,\tau,\mathcal{C}) \in \mathbb{R}^{N\times H\times 3}$. We
optimize the motion-weighted flow-matching objective
\begin{equation}
\mathcal{L} \;=\;
\mathbb{E}_{\tau,F_0}\!\left[\,\sum_{i,t} w_{i,t}\,
\bigl\lVert u_\theta(F_\tau,\tau,\mathcal{C})_{i,t} - u^{\star}_{i,t} \bigr\rVert_2^{2}\right],
\end{equation}
where
\(w_{i,t} = m_{i,t}\cdot\max\!\bigl(
\sigma\!\bigl((\lVert\Delta q_{i,t}\rVert - \tau_w)\,\kappa/\tau_w\bigr)^{\gamma},
\, w_{\min}\bigr)\)
combines the trajectory-validity mask $m_{i,t}$ with a soft motion-saliency term. Here $m_{i,t}$ excludes tracks affected by occlusion, invalid depth, or SpatialTrackerV2 confidence below $0.05$; $\tau_w$ is the motion-magnitude threshold; $\kappa$ controls the sigmoid sharpness; $\gamma$ controls the saliency exponent; and $w_{\min}$ preserves a nonzero loss on static points. At inference, $\mathcal{F}$ is recovered from
$F_0\sim\mathcal{N}(0,I)$ by an explicit Euler integration of $u_\theta$ over a
small number of flow steps and then mapped back to metric displacements using the inverse of the training normalization.

\begin{table}[t]
\centering
\caption{\textbf{Model configurations across simulation benchmarks and real-world experiments.} The 3D action denoiser shared attention layers are the action--scene fusion blocks that are shared by translation and rotation prediction. Rotation format is the policy rotation representation, with RoboTwin2.0 rotations converted to 6D internally; action horizon is the number of future actions predicted per policy query, and embodiment specifies the controlled arm setup.}
\label{tab:model_config}
\resizebox{\linewidth}{!}{
\begin{tabular}{llcccc}
\toprule
\multicolumn{2}{c}{\textbf{Module}} 
& \textbf{CALVIN} 
& \textbf{LIBERO} 
& \textbf{RoboTwin2.0 -- five-task ablations}
& \textbf{Real-world}\\
\midrule

\multirow{8}{*}{\textbf{Trajectory Denoiser}}
& Hidden dim      & 512 & 384 & 320 & 128 \\
& Encoder layers         & 6 & 6 & 6 & 4 \\
& Decoder layers        & 8 & 8 & 8 & 8 \\
& Attention heads        & 16 & 8 & 8 & 8 \\
& Batch size      & 128 & 128 & 128 & 128 \\
& Optimizer       & AdamW & AdamW & AdamW & AdamW \\
& Learning rate   & 1e-4 & 1e-4 & 1e-4 & 1e-4 \\
& Iterations      & 20k & 20k & 20k & 20k \\
\midrule
\multirow{10}{*}{\textbf{Action Denoiser}}
& Hidden dim      & 960 & 912 & 912 & 256 \\
& Shared attention layers   & 16 & 16 & 16 & 16 \\
& Attention heads        & 16 & 16 & 16 & 16 \\
& Batch size      & 32 & 128 & 64 & 32 \\
& Optimizer       & AdamW & AdamW & AdamW & AdamW \\
& Learning rate   & 2e-5 & 5e-5 & 4e-5 & 2e-5 \\
& Iterations      & 50k & 80k & 300k & 50k \\
& Rotation format & Euler & Axis-angle & 6D rotation & 6D rotation \\
& Action horizon  & 10 & 10 & 16 & 8 \\
& Embodiment      & Single-arm & Single-arm & Bimanual & Single-arm \\
\bottomrule
\end{tabular}
}
\end{table}

\paragraph{RoboTwin2.0 training protocol and full results.}
We train one multi-task policy on all 50 RoboTwin2.0 tasks, using 50 clean and 500 domain-randomized demonstrations per task (27,500 demonstrations in total). We follow the official RoboTwin2.0 data-collection and Easy/Hard evaluation protocols. We fine-tune all $\pi_{0.5}$ parameters from its public base checkpoint. \model{} uses a pretrained VLM backbone but no robot-action pretraining and is optimized only on these RoboTwin2.0 demonstrations. Both policies predict end-effector pose actions. Table~\ref{tab:robotwin2_all} reports per-task success rates under the Easy and Hard randomization settings.

\begin{table*}[t]
\centering
\caption{\textbf{Per-task success rates on 50 RoboTwin2.0 tasks.}
Each task is evaluated under both \texttt{Easy} and \texttt{Hard} domain
randomization, with 50 rollouts per setting. We report success rates as percentages.
\textbf{Average: $\boldsymbol{\pi_{0.5}}$ Easy 75.9 / Hard 75.7, GeomVLA
Easy 78.6 / Hard 76.2.}}
\label{tab:robotwin2_all}
\small
\setlength{\tabcolsep}{3.8pt}
\renewcommand{\arraystretch}{1.03}
\begin{tabular}{@{}lrrrr@{\hspace{1.2em}}lrrrr@{}}
\toprule
\multicolumn{1}{c}{\multirow{2}{*}[-0.3ex]{\textbf{Task}}}
& \multicolumn{2}{c}{$\boldsymbol{\pi_{0.5}}$}
& \multicolumn{2}{c}{\textbf{GeomVLA}}
& \multicolumn{1}{c}{\multirow{2}{*}[-0.3ex]{\textbf{Task}}}
& \multicolumn{2}{c}{$\boldsymbol{\pi_{0.5}}$}
& \multicolumn{2}{c}{\textbf{GeomVLA}} \\
\cmidrule(lr){2-3} \cmidrule(lr){4-5}
\cmidrule(lr){7-8} \cmidrule(l){9-10}
& \textbf{Easy} & \textbf{Hard} & \textbf{Easy} & \textbf{Hard}
& & \textbf{Easy} & \textbf{Hard} & \textbf{Easy} & \textbf{Hard} \\
\midrule
Adjust Bottle         & 86 & 92 & 100 & 100
& Place Can Basket      & 70 & 70 & 84 & 76 \\
Beat Block Hammer     & 84 & 80 & 88 & 90
& Place Cans Plasticbox & 74 & 86 & 94 & 92 \\
Blocks Ranking RGB    & 88 & 84 & 46 & 44
& Place Container Plate & 96 & 98 & 100 & 98 \\
Blocks Ranking Size   & 52 & 38 & 24 & 32
& Place Dual Shoes      & 62 & 64 & 84 & 86 \\
Click Alarmclock      & 94 & 98 & 98 & 96
& Place Empty Cup       & 90 & 90 & 100 & 96 \\
Click Bell            & 98 & 98 & 100 & 100
& Place Fan             & 88 & 84 & 88 & 86 \\
Dump Bin Bigbin       & 84 & 90 & 72 & 68
& Place Mouse Pad       & 44 & 36 & 88 & 90 \\
Grab Roller           & 98 & 96 & 100 & 100
& Place Object Basket   & 70 & 80 & 48 & 56 \\
Handover Block        & 42 & 32 & 14 & 10
& Place Object Scale    & 80 & 70 & 96 & 98 \\
Handover Mic          & 42 & 52 & 72 & 38
& Place Object Stand    & 86 & 94 & 96 & 98 \\
Hanging Mug           & 16 & 14 & 6 & 4
& Place Phone Stand     & 64 & 66 & 76 & 84 \\
Lift Pot              & 88 & 88 & 32 & 18
& Place Shoe            & 88 & 84 & 96 & 100 \\
Move Can Pot          & 92 & 88 & 100 & 98
& Press Stapler         & 70 & 78 & 70 & 78 \\
Move Pillbottle Pad   & 80 & 88 & 96 & 94
& Put Bottles Dustbin   & 34 & 38 & 34 & 18 \\
Move Playingcard Away & 74 & 76 & 98 & 96
& Put Object Cabinet    & 56 & 60 & 62 & 48 \\
Move Stapler Pad      & 56 & 64 & 72 & 76
& Rotate QRcode         & 96 & 84 & 80 & 80 \\
Open Laptop           & 90 & 92 & 94 & 92
& Scan Object           & 62 & 56 & 62 & 60 \\
Open Microwave        & 82 & 80 & 94 & 82
& Shake Horizontally    & 96 & 98 & 100 & 100 \\
Pick Diverse Bottles  & 86 & 84 & 86 & 84
& Shake Bottle          & 96 & 98 & 100 & 100 \\
Pick Dual Bottles     & 94 & 94 & 100 & 96
& Stack Blocks Three    & 56 & 60 & 46 & 32 \\
Place A2B Left        & 84 & 90 & 90 & 90
& Stack Blocks Two      & 92 & 92 & 74 & 88 \\
Place A2B Right       & 82 & 72 & 96 & 90
& Stack Bowls Three     & 78 & 64 & 46 & 36 \\
Place Bread Basket    & 74 & 68 & 86 & 86
& Stack Bowls Two       & 88 & 98 & 84 & 82 \\
Place Bread Skillet   & 72 & 66 & 92 & 80
& Stamp Seal            & 88 & 74 & 96 & 96 \\
Place Burger Fries    & 78 & 82 & 96 & 94
& Turn Switch           & 56 & 58 & 72 & 72 \\
\bottomrule
\end{tabular}
\end{table*}

\paragraph{Two-seed stability check on CALVIN and LIBERO.}
We train one additional \model{} model on each benchmark. The CALVIN scores are 4.624 and 4.609 for seeds 0 and 42, and the LIBERO success rates are 98.4\% and 98.0\%. These differences are smaller than the principal ablation gaps in Table~\ref{tab:ablation_results}.

\paragraph{Evaluation of the 3D scene trajectory prediction horizon.}
We examine how the prediction horizon affects trajectory accuracy and downstream policy performance. Keeping all other components fixed, we train \model{} on the same CALVIN demonstrations using either the default horizon $H=15$ or a longer horizon $H=32$, and evaluate both trajectory predictors on held-out valid moving-point tracks over their shared first 15 predicted steps. SpatialTrackerV2 provides the pseudo-reference tracks. As shown in Table~\ref{tab:horizon_ablation}, increasing the horizon from $H=15$ to $H=32$ increases the mean error from 2.46\,cm to 3.35\,cm and the endpoint error from 4.61\,cm to 6.43\,cm. The CALVIN score also decreases from 4.624 to 4.272, suggesting that requiring the denoiser to predict farther into the future may make the trajectory-learning problem harder and weaken the motion representation used by the downstream policy.

\begin{table}[t]
    \centering
    \small
    \setlength{\tabcolsep}{7pt}
    \caption{\textbf{Effect of trajectory-prediction horizon on CALVIN.} Errors are computed on held-out valid moving points against SpatialTrackerV2 pseudo-reference tracks over the first 15 future steps for both models. In this two-setting comparison, the longer training horizon yields higher trajectory error and lower downstream performance.}
    \label{tab:horizon_ablation}
    \begin{tabular}{cccc}
        \toprule
        Horizon
        & Mean error (cm) $\downarrow$
        & Endpoint error (cm) $\downarrow$
        & Perf. $\uparrow$ \\
        \midrule
        $H=15$ & \textbf{2.46} & \textbf{4.61} & \textbf{4.624} \\
        $H=32$ & 3.35          & 6.43          & 4.272 \\
        \bottomrule
    \end{tabular}
\end{table}

\paragraph{Instruction conditioning for the 3D Scene Trajectory Denoiser.}

We examine whether the predicted trajectories capture instruction-conditioned motion rather than generic scene dynamics. We replace the correct task instruction with a mismatched instruction while keeping the visual observation unchanged. Because changing only the instruction changes the trajectory prediction and increases its error relative to the matched-task pseudo-reference, the denoiser is sensitive to language conditioning.

\begin{table}[t]
    \centering
    \small
    \setlength{\tabcolsep}{6pt}
    \caption{\textbf{Instruction sensitivity of predicted 3D scene trajectories.} Endpoint error is measured in centimeters on held-out valid moving points against SpatialTrackerV2 pseudo-reference tracks at $H=15$. Replacing the matched instruction with a mismatched instruction increases the error on all three benchmarks.}
    \label{tab:traj_instruction}
    \begin{tabular}{lccc}
        \toprule
        Benchmark
        & Matched $\downarrow$
        & Mismatched $\downarrow$
        & Increase \\
        \midrule
        CALVIN      & \textbf{4.61} & 8.01  & $+3.40$ \\
        LIBERO      & \textbf{9.12} & 14.56 & $+5.44$ \\
        RoboTwin2.0 & \textbf{5.77} & 7.41  & $+1.64$ \\
        \bottomrule
    \end{tabular}
\end{table}

\paragraph{Control for model capacity.}
We replace the learned 3D Scene Trajectory Denoiser with a randomly initialized module of the same size, freeze it, and retrain the policy. The CALVIN score decreases from 4.624 to 4.443, indicating that the gain is associated with learned motion features rather than the added parameter count alone.

\paragraph{Inference latency.}
We measure the latency of one \texttt{model.step()} call during online CALVIN evaluation; each call predicts a chunk of 10 actions, all of which are executed before the next policy query. Measurements use CUDA-synchronized timers during PyBullet rollouts with EGL rendering on one NVIDIA L40S GPU. For each model, we evaluate four instruction sequences and discard the first 10 chunks as warm-up. \model{} without motion requires $219.0 \pm 11.3$ ms per chunk over 165 calls: 35.5 ms for the VLM encoder and 182.4 ms for the 3D action policy. Full \model{} requires $280.0 \pm 14.0$ ms over 155 calls: 37.2 ms for the VLM encoder, 37.5 ms for the trajectory field, and 202.1 ms for the action policy. Full \model{} therefore adds 61.0 ms per chunk, or 27.9\% overhead. The trajectory field accounts for 61.5\% of this increase, while most of the remainder arises from the slower action-policy stage. In both models, the action policy dominates total inference time.

\paragraph{Ablation variant descriptions.}
Table~\ref{tab:ablation_descriptions} provides detailed descriptions of the
model variants used in the main-paper ablation study. These variants isolate the effects of 3D future-motion
reasoning, 3D action denoising, motion-token extraction, and coordinate-frame
alignment.

\begin{table*}[t]
\centering
\small
\setlength{\tabcolsep}{4pt}
\caption{\textbf{Ablation model descriptions.}
Detailed descriptions of the model variants evaluated in the ablation study.
Each variant modifies one aspect of \model{}, including future-motion reasoning,
action denoising, motion-token extraction, or coordinate-frame alignment, while
the corresponding quantitative results are reported in the main paper.}
\begin{tabularx}{\textwidth}{@{}p{0.30\textwidth}X@{}}
\toprule
\textbf{Variant} & \textbf{Description} \\
\midrule
\model{} w/o motion
& Removes the 3D Scene Trajectory Denoiser and directly predicts actions with the 3D action denoiser. \\
2DGeomVLA w/o motion
& Uses a 2D image-based action denoiser without future-motion reasoning. \\
2DGeomVLA
& Uses 2D future-motion reasoning with a 2D action denoiser. \\
2DGeomVLA LaMP-style
& Uses image-plane $(u,v)$ coordinates and metric depth for motion reasoning instead of metric 3D points. \\
\model{} w/ 2D motion
& Uses image-space future motion with the 3D action denoiser. \\
\model{} w/ denoised trace
& Conditions action denoising on the fully denoised 3D trajectory. \\
\model{} w/ denoised latent
& Conditions action denoising on the fully denoised motion latent. \\
\model{} w/ partial denoising
& Reads the motion latent at $\tau = 0.1$ instead of pure noise. \\
\model{} encoder-only
& Uses only the trajectory-denoiser encoder, without velocity prediction. \\
\model{} w/ arbitrary frame
& Expresses scene, motion, and action in a shared arbitrary rigid frame. \\
\model{}
& Unifies scene, future motion, and action in a shared 3D frame. \\
\bottomrule
\end{tabularx}
\label{tab:ablation_descriptions}
\end{table*}

\paragraph{Real-world task descriptions.}
We provide detailed descriptions of all eight real-world manipulation tasks, grouped by the primary capability they evaluate. Together, these tasks assess precise pose control, fine-grained small-object manipulation, color grounding, and multi-stage spatial reasoning. The number of demonstrations collected for each task is reported in parentheses.

\vspace{0.3em}
\noindent\textit{Precision and orientation control.} 
\begin{itemize} 
\item \textbf{Insert marker} (50 demos). The robot picks up a marker lying flat on the table and inserts it vertically into a cup. This task requires accurate geometric reasoning and precise end-effector orientation under tight spatial tolerances.
\item \textbf{Stand bottle upright} (20 demos). The robot picks up a small bottle lying flat on the table, rotates it to an upright orientation, and places it stably on the table. The task requires a large end-effector rotation followed by controlled placement to prevent the bottle from falling.
\item \textbf{Transfer marker} (20 demos). The robot picks up a marker standing inside one cup and inserts it into another. Because the marker pose varies across trials, the policy must adapt its gripper orientation for both grasping and insertion. It must also avoid lifting or knocking over the source cup while extracting the marker. \end{itemize}

\vspace{0.3em} \noindent\textit{Fine-grained small-object manipulation.} 
\begin{itemize} 
\item \textbf{Stack blocks} (50 demos). The robot picks up a small block and stacks it on another block of identical size and shape. Successful execution requires accurate grasping, close alignment, and a controlled low-height release; otherwise, the upper block can easily slide off.
\item \textbf{Uncap marker} (20 demos). The robot identifies and grasps the cap of a marker lying flat on the table and pulls it off. This task requires the policy to distinguish the cap from the marker body and precisely manipulate only the cap. 
\end{itemize}

\vspace{0.3em} \noindent\textit{Color grounding and multi-stage spatial reasoning.} 
\begin{itemize} 
\item \textbf{Place bottles} (50 demos). The robot places the green bottle into the green cup and then places the blue bottle into the blue cup. This two-stage task evaluates color grounding and sequential object placement.
\item \textbf{Rank by color} (20 demos). The robot sequentially arranges three bottles from left to right, as viewed from the wrist camera, in pink--green--blue order. This three-stage, long-horizon task combines color grounding, ordered spatial reasoning, and sequential manipulation.
\item \textbf{Place in triangle} (20 demos). Using the middle bottle as an anchor vertex, the robot sequentially moves the left and right bottles, as viewed from the wrist camera, so that the three bottles form a triangle. This two-stage task evaluates reasoning about relative spatial relationships and precise placement. 
\end{itemize}

\begin{figure*}[t!] 
\centering 
\includegraphics[width=\linewidth]{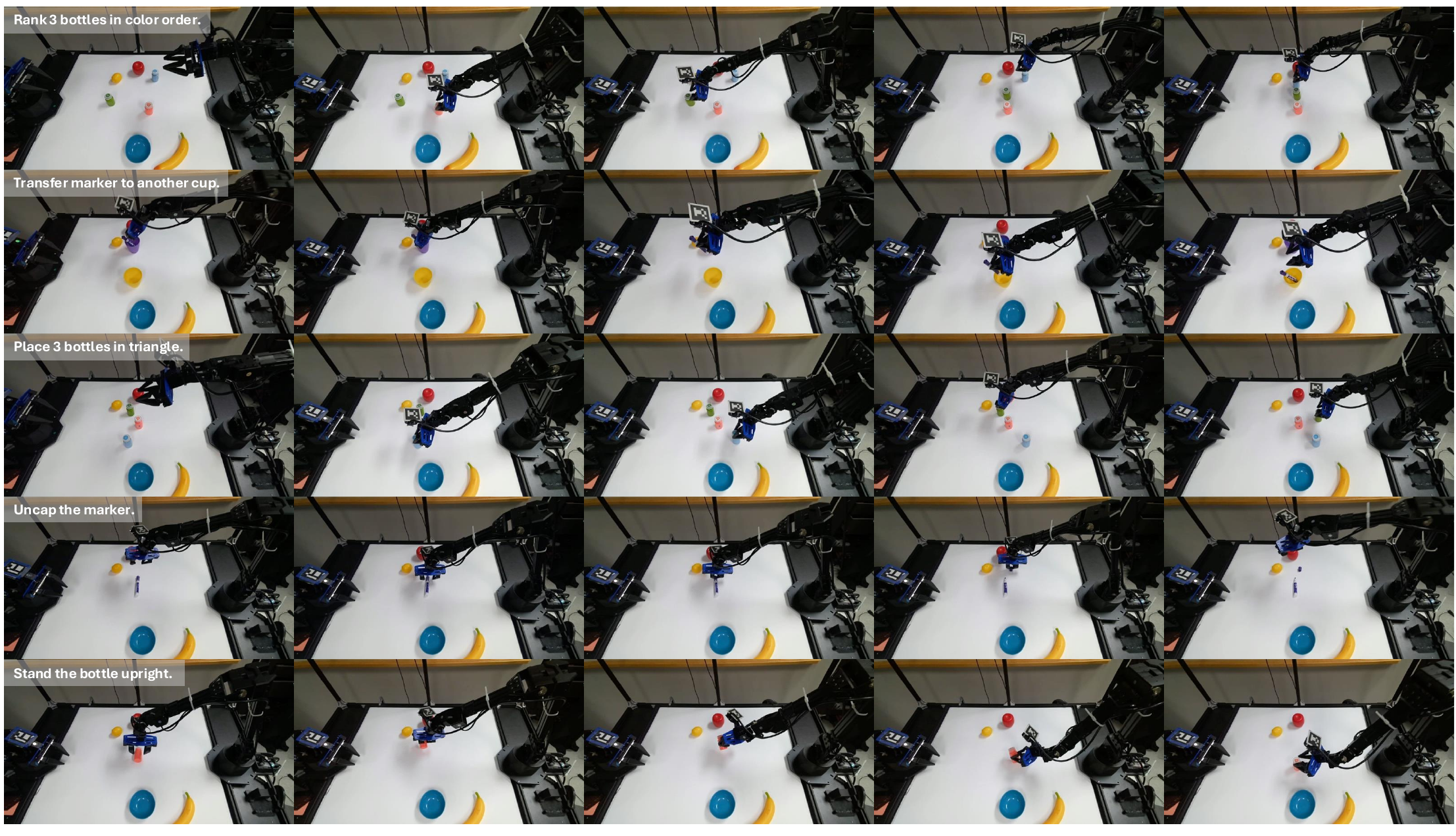} 
\caption{\textbf{Qualitative \model{} rollouts on five additional real-world tasks.} These tasks complement those shown in Figure~\ref{fig:real-world}. 
Each row shows a temporal sequence from one successful rollout. From top to bottom, the tasks are Rank by color, Transfer marker, Place in triangle, Uncap marker, and Stand bottle upright.} \label{fig:realworld_rollouts} \end{figure*}

During evaluation, we vary distractors, goal objects, and object layouts across both seen and unseen settings. Figure~\ref{fig:realworld_rollouts} shows representative successful \model{} rollouts on the five real-world tasks trained with 20 demonstrations each. Each row presents a temporal sequence from one rollout, illustrating how the policy handles diverse challenges such as precise object reorientation, small-object grasping, partial occlusion, and multi-stage spatial reasoning. These qualitative examples complement the quantitative results by showing the execution process behind successful task completion.

\end{document}